\documentclass{article}

\PassOptionsToPackage{numbers, sort&compress}{natbib}

\usepackage[preprint]{neurips_2024}

\usepackage[utf8]{inputenc} 
\usepackage[T1]{fontenc}    
\usepackage{hyperref}       
\usepackage{url}            
\usepackage{booktabs}       
\usepackage{amsfonts}       
\usepackage{nicefrac}       
\usepackage{microtype}      
\usepackage{xcolor}         
\usepackage{subcaption}
\usepackage{graphicx}
\usepackage{booktabs}
\usepackage{soul}
\usepackage{pifont}
\usepackage{verbatim}
\usepackage{multirow}
\usepackage{caption}
\usepackage{amsmath}
\usepackage{colortbl}
\usepackage{booktabs}

\usepackage{array}

\newcommand\flops{FLOPs}

\title{Expressive Keypoints for Skeleton-based Action Recognition via Skeleton Transformation}

\author{%
  Yijie Yang$^1$\thanks{Both authors contributed equally.}, 
  \And
  Jinlu Zhang$^{2,*}$, 
  \And
  Jiaxu Zhang$^1$, 
  \And
  Zhigang Tu$^1$\thanks{Corresponding author: tuzhigang@whu.edu.cn} \\
  \And
  \normalfont{$^1$Wuhan University, $^2$Peking University}\\
  \texttt{\{yijieyang, zjiaxu, tuzhigang\}@whu.edu.cn, jinluzhang@pku.edu.cn}
}

\begin{document}

\maketitle

\begin{abstract}

In the realm of skeleton-based action recognition, the traditional methods which rely on coarse body keypoints fall short of capturing subtle human actions. In this work, we propose Expressive Keypoints that incorporates hand and foot details to form a fine-grained skeletal representation, improving the discriminative ability for existing models in discerning intricate actions. To efficiently model Expressive Keypoints, the Skeleton Transformation strategy is presented to gradually downsample the keypoints and prioritize prominent joints by allocating the importance weights. Additionally, a plug-and-play Instance Pooling module is exploited to extend our approach to multi-person scenarios without surging computation costs. Extensive experimental results over seven datasets present the superiority of our method compared to the state-of-the-art for skeleton-based human action recognition. Code is available at \url{https://github.com/YijieYang23/SkeleT-GCN}.

\end{abstract}
\vspace{-2mm}
\section{Introduction}
\label{sec:introduction}


Skeleton-based action recognition has become a cornerstone for numerous vision applications such as video surveillance~\cite{surveillance1,surveillance2}, human-robot interaction~\cite{human-robot}, and sports analytics~\cite{sports}, due to its succinct representation and robustness to variations in lighting, scale, and viewpoint. 
Traditional methods primarily utilize simple body keypoints defined in NTU~\cite{ntu60,ntu120} and COCO~\cite{mscoco} formats, to provide sparse representations of human motion. Despite their utility, the over concise representations are constrained by missing subtle but critical details involving hand and foot movements. Consequently, existing coarse skeletal representations are limited in effectively distinguishing intricate actions.

\begin{figure*}[t]
    \centering
    \begin{subfigure}{0.55\textwidth}
        \centering
        \includegraphics[width=\textwidth]{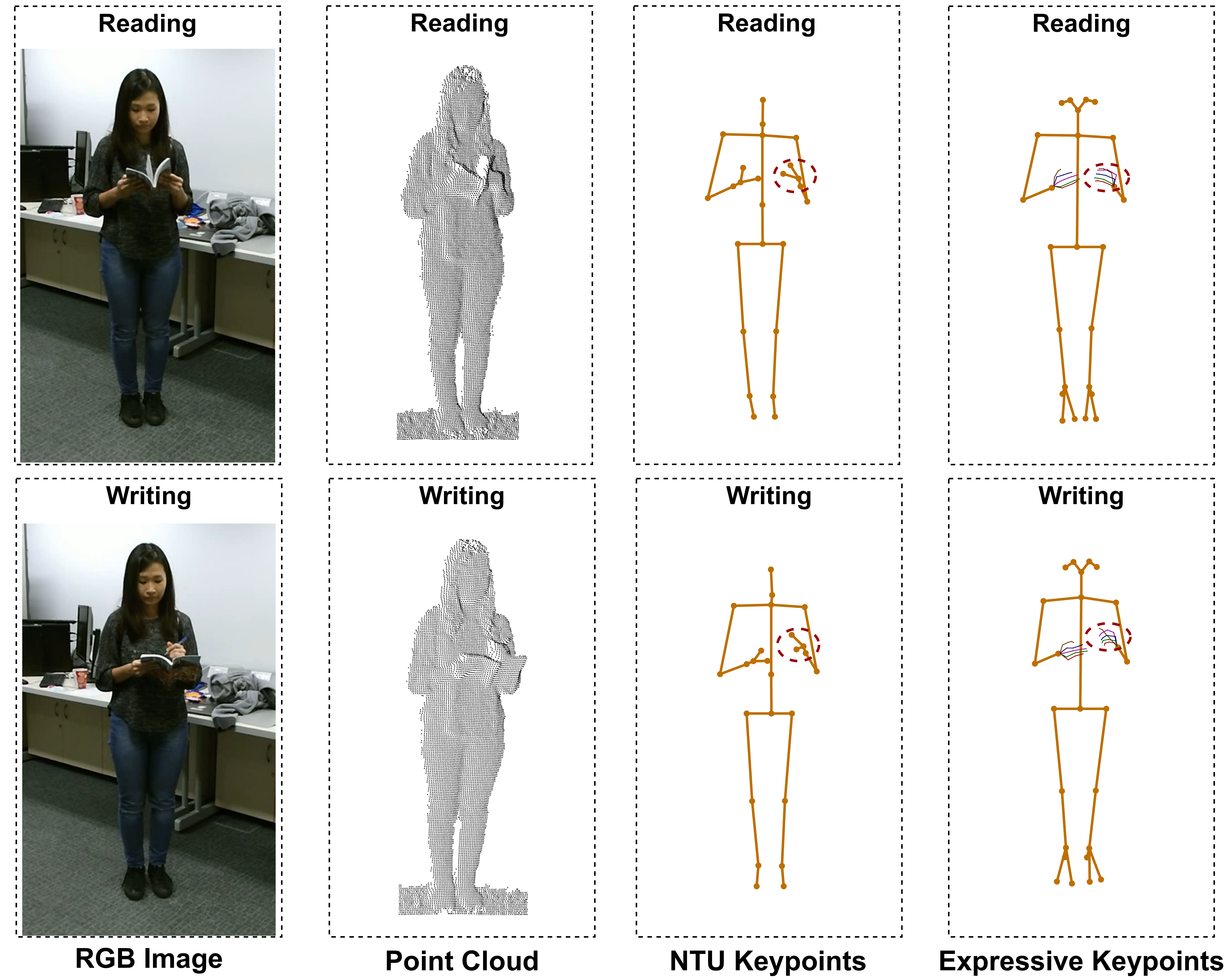}
        \caption{Various representations of the actions \textit{Reading} and \textit{Writing}.}
        \label{fig:teaser-a}
    \end{subfigure}
    \hfill
    \begin{subfigure}{0.42\textwidth}
        \centering
        \includegraphics[width=\textwidth]{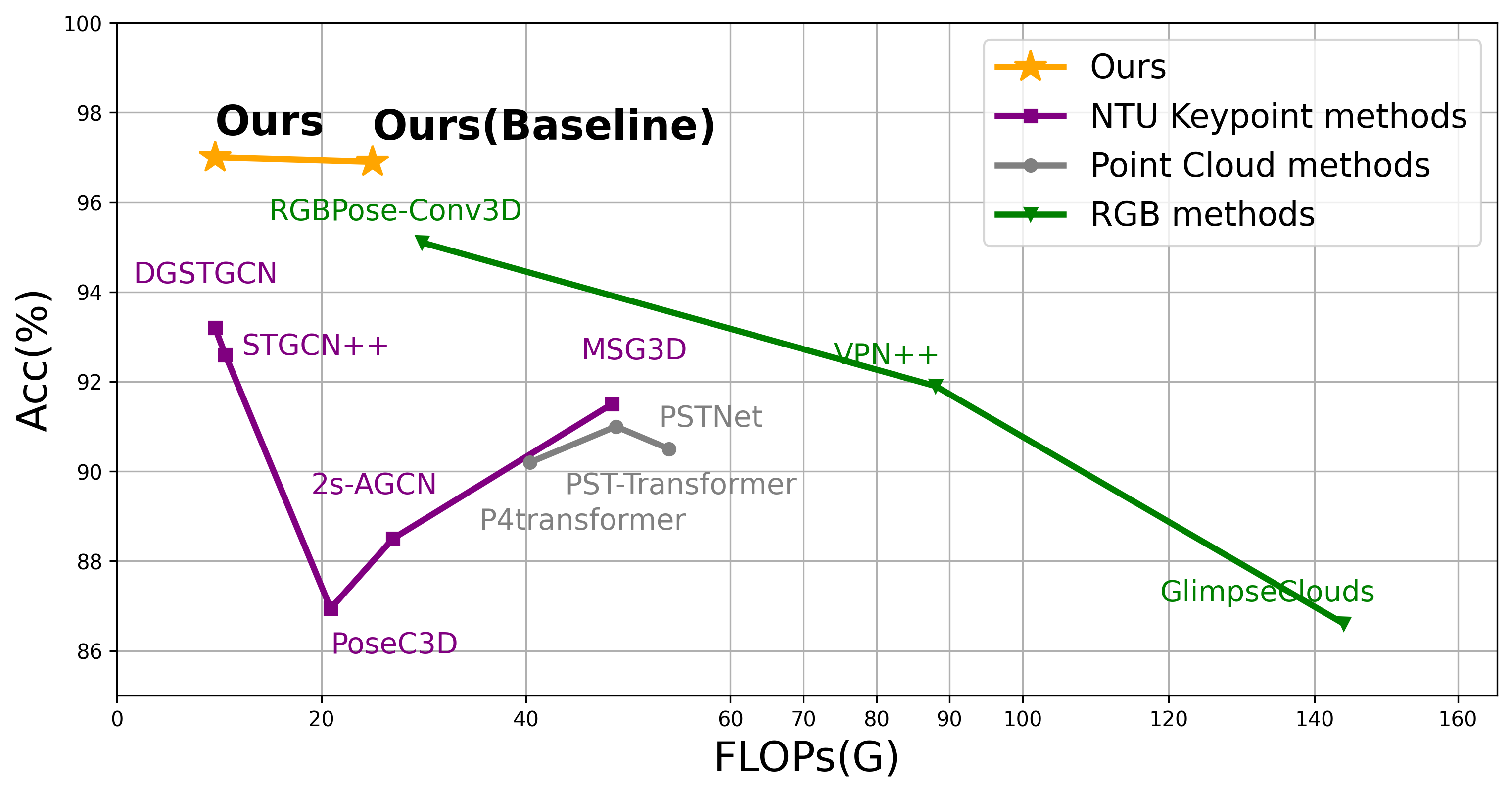}
        \includegraphics[width=\textwidth]{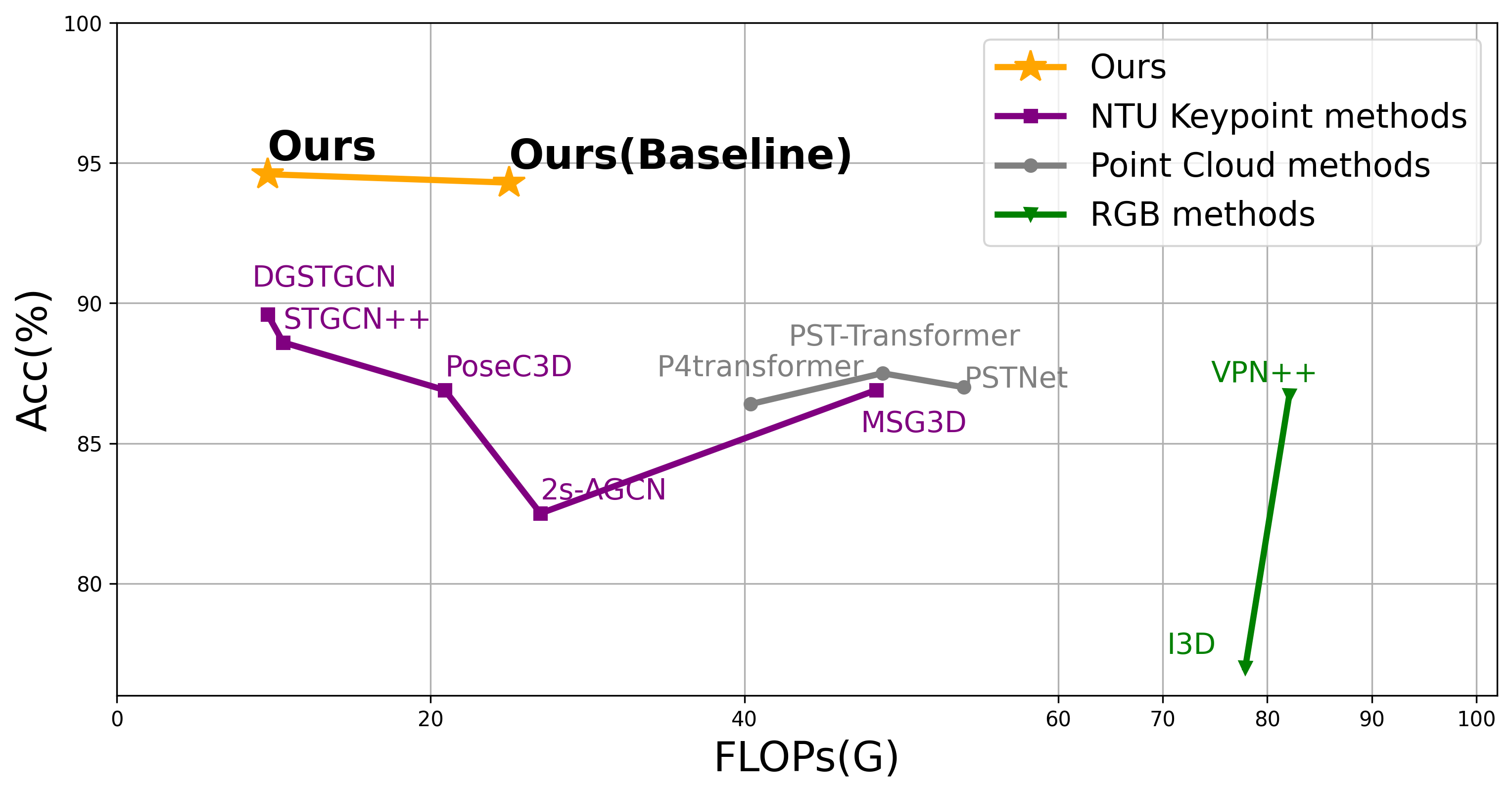}
        \caption{Accuracy and efficiency comparison.}
        \label{fig:teaser-b}
    \end{subfigure}
    \caption{\textbf{(a).} Various representations of the same actions. \textbf{(b).} Accuracy and efficiency comparison of our method and the representative methods on NTU-60~\cite{ntu60} (Top) and NTU-120~\cite{ntu120} (Bottom).}
    \label{fig:teaser}
    \vspace{-5mm}
\end{figure*}

Recently, some approaches~\cite{P4Transformer,PSTNet,PST-Transformer} have resorted to the point cloud representation to capture the detailed spatial structure of human surface, thereby enhancing the ability to recognize complex movements. However, it comes with enormously increased computational cost, detracting from the efficiency of point-based representation. Moreover, several studies~\cite{OHA-GCN,JLF,skp} have aimed to improve the recognition accuracy by introducing object points. However, the generalization of these methods is limited especially in the human-centric scenarios where no interacted object involved.



To solve the limitations of prior works, we incorporate richer limb keypoints into body keypoints to propose a fine-grained representation called Expressive Keypoints. It emphasizes nuanced hand interactions and foot movements which are crucial to discerning subtle actions. As shown in ~Fig.~\ref{fig:teaser-a}, we present various data representations that are commonly utilized. 
Compared to the representations of RGB images, excessive point cloud data, and coarse body keypoints, the Expressive Keypoints representation stands out for its insensitivity to viewpoints, relatively small data footprint, and ability to represent fine-grained limb details. In practice, Expressive Keypoints can be easily estimated from RGB images based on COCO-Wholebody~\cite{cocowholebody} annotations, without relying on obtaining depth information from multi-view data or lab-controlled motion capture system. Experimental results demonstrate that all three baseline methods~\cite{PYSKL,CTRGCN,DGSTGCN} achieve significant improvement in accuracy (+ over 6\%) when replacing coarse-grained keypoints with Expressive Keypoints.


However, the computational cost of directly taking Expressive Keypoints as input also scales considerably, since nearly three times more joints need to be dealt with. To enhance the computationally efficiency, we propose the Skeleton Transformation (SkeleT) strategy to gradually downsample the skeletal representation of Expressive Keypoints across multiple stages. This novel strategy involves the learnable mapping matrices to refine skeleton features by re-weighting and downsampling the keypoints. These mapping matrices are initialized by semantic partitioning of human topology, and iteratively optimized during training. By further introducing variable group design for different skeletal scales, skeleton features are evenly split and transformed independently before concatenation. SkeleT strategy enables effective downsampling of keypoints and nuanced modeling in groups. It can be effortlessly integrated into most existing GCN-based skeleton action recognition methods, forming our SkeleT-GCN to efficiently process Expressive Keypoints. In experiments over four standard skeleton action recognition datasets~\cite{ntu60,ntu120,pku,ucla}, SkeleT-GCN achieves comparable or even higher accuracy with much lower (less than half) \flops ~compared to its baseline GCN method.

Moreover, we want to further evaluate our method on the general in-the-wild datasets~\cite{k400,ucf101,hmdb} which include multi-person group activity scenarios. However, we find that traditional GCN methods perform feature modelling for each input person individually and conduct feature fusion in the late stage. Consequently, they have the limitation of exponentially increasing computational complexity as the number of individuals grows in a wild scene. Inspired by~\cite{skp}, we implement a lightweight Instance Pooling module before the GCN models. The key idea is to aggregate the features of multiple persons and projects them to a single skeletal representation in the early stage. By exploiting the plug-and-play Instance Pooling module, the classification of group activities can be supported without surging computation cost. This offers a practical and viable solution for extending GCN-based skeleton action recognition methods (including our SkeleT-GCN) to multi-person scenarios.

In extensive experimental evaluations over the total of seven datasets~\cite{ntu60,ntu120,pku,ucla,k400,ucf101,hmdb}, our pipeline consistently achieves the state-of-the-art across all the benchmarks (see Fig.~\ref{fig:teaser-b}), demonstrating its superior performance and robust generalization. We find that strategically employing fine-grained keypoints enables recognizing intricate human actions with efficient computation complexity. In summary, the main contributions of our work are threefold:
\begin{itemize}
\item We introduce fine-grained limb details as the Expressive Keypoints representation for skeleton action recognition,  boosting the performance in identifying intricate actions.
\item We propose the Skeleton Transformation strategy to make existing GCN methods highly efficient while preserving accuracy, through dynamically downsampling of keypoints.
\item We implement a plug-and-play Instance Pooling module to extend GCN methods to multi-person group activity scenarios without surging computation cost.
\end{itemize}

\section{Related works}
\label{sec:related_works}



\subsection{Point-based action recognition}
Point-based action recognition methods are more robust against variations of lightning and view variation compared with RGB-based methods~\cite{2-stream,I3D,X3D,slowfast}. 
Some works~\cite{pointnet++,P4Transformer,PSTNet,PST-Transformer} take point cloud data, which consists of numerous unordered 3D point sets, as input for their methods. However, point cloud data introduces too much redundant information for learning action patterns, leading to high computation costs.
Some works utilize 2D/3D keypoints~\cite{ntu60,mscoco} to represent the skeletal structure of human body. They are also commonly referred to as skeleton-based methods. Among them, 
GCN models~\cite{STGCN,MSAAGCN,CTRGCN,PYSKL,DGSTGCN,InfoGCN,HDGCN} have been adopted frequently due to the effective representation for the graph structure~\cite{survey}. Additionally, some models~\cite{PoseC3D,SkeleMotion,skeleton-image} attempt to project human body keypoints into multiple 2D pseudo-images to learn useful features, which also achieves notable performance. 
Nevertheless, existing skeleton-based methods use coarse-grained skeleton representation as input, leading to the challenge of discerning complex actions, which results in limited performance. 
To this end, we propose to incorporate hand and foot keypoints into the body part, forming a fine-grained skeletal structure to better distinguish the intricate actions.

\subsection{GCNs for skeleton-based action recognition}

STGCN~\cite{STGCN} first utilized graph convolution to conduct skeleton action recognition, GCN-based methods soon became the mainstream. Different improvements have been made in recent works~\cite{MSAAGCN,CTRGCN,PYSKL,DGSTGCN}. MS-AAGCN~\cite{MSAAGCN} proposes to adaptively learn the topology of graphs instead of setting it manually. 
CTRGCN~\cite{CTRGCN} takes a shared topology matrix as the generic prior for network channels to improve performance. PYSKL~\cite{PYSKL} presents an open-source toolbox for skeleton-based action recognition, which benchmarked representative GCN methods with good practices. DGSTGCN~\cite{DGSTGCN} proposes a lightweight yet powerful model without a predefined graph.
However, traditional methods commonly face two limitations: (1) they maintain a static skeleton structure with a fixed number of keypoints, which restricts their ability to capture multi-scale information, and (2) the computational costs linearly increase with each additional person, resulting in the input being cropped to a maximum of two individuals. In this work, we propose a Skeleton Transformation strategy to dynamically modify the skeleton structure and downsample keypoints. Additionally, we introduce an Instance Pooling module to overcome the constraints of input individuals.


% ...

\begin{figure*}[thp]
    \centering
    \includegraphics[width=0.99\textwidth]{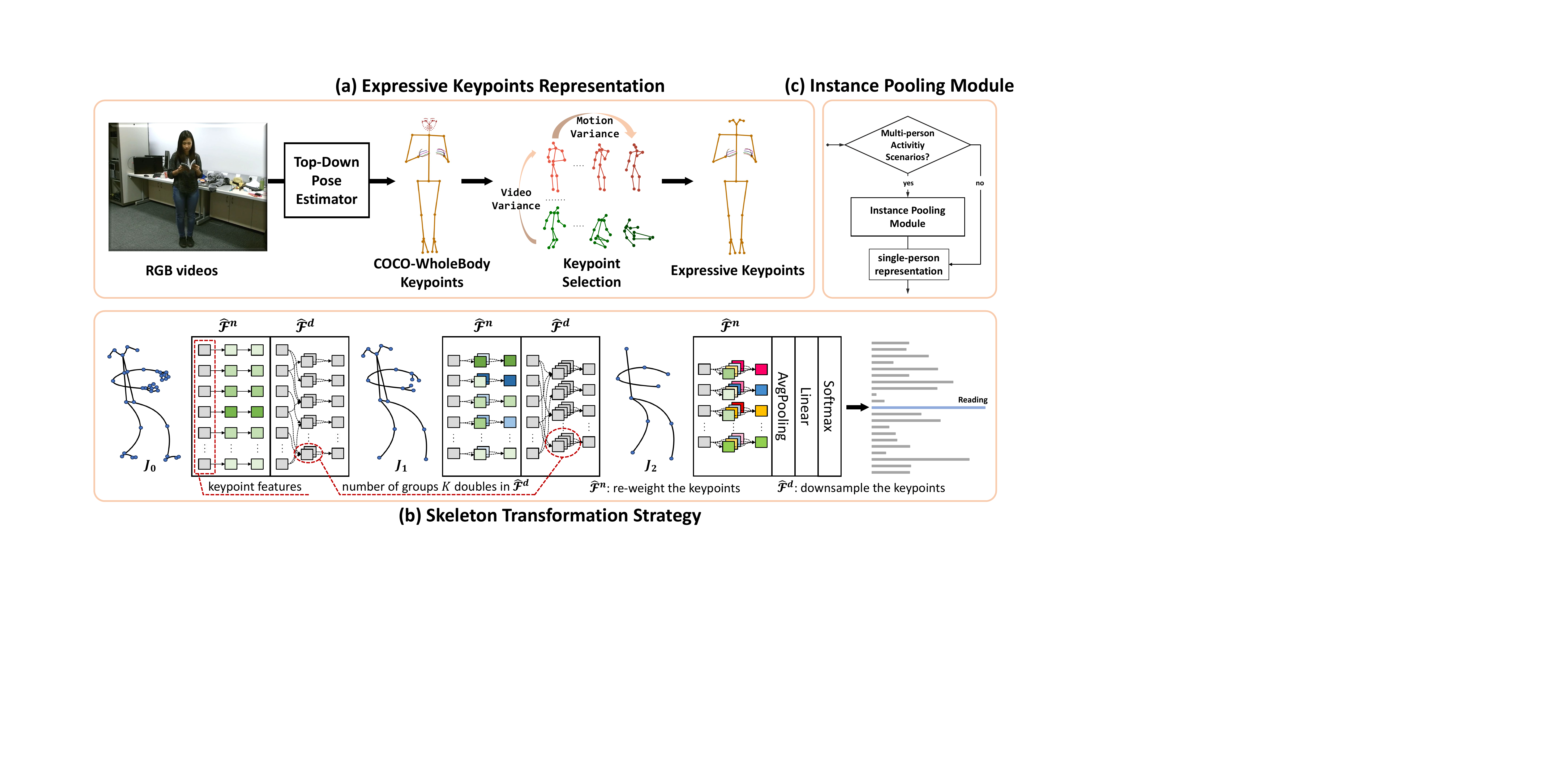}
    \caption{\textbf{Overview of proposed pipeline. (a).} We use a top-down estimator to extract COCO-WholeBody Keypoints from videos, and conduct keypoint selection based on statistical metrics to remove the redundant facial keypoints, forming our \textbf{Expressive Keypoints representation}. \textbf{(b).} We propose the \textbf{Skeleton Transformation strategy} that can be integrated into most GCN methods to efficiently process Expressive Keypoints. It guides the network to alter the skeletal features in groups by re-weighting and gradually downsampling the keypoints. \textbf{(c).} We implement a \textbf{Instance Pooling module} to fuse the multiple instances in the early stage. We use it as an lightweight extension for evalutaion our methods in general wild scenarios, which contains multi-person group activities.}
    \label{fig:framework}
    \vspace{-1mm}
\end{figure*}

\section{Proposed pipeline}
\label{sec:pipeline}
The overview of our proposed pipeline is depicted in~Fig.~\ref{fig:framework}. In Sec.~\ref{sec:expressive-keypoints}, we incorporate detailed keypoints of limbs to coarse-grained body keypoints, forming the representation of Expressive Keypoints. We elaborate on the collection and preprocessing of these keypoints, highlighting the benefits of this approach.  In Sec.~\ref{sec:gts}, we propose the Skeleton Transformation strategy to efficiently deal with more limb keypoints. We find that implicitly aggregating keypoint in latent space in the network processing can significantly reduce computational complexity while maintaining high accuracy. In Sec.~\ref{sec:ip-module}, we discover that individual modeling and late fusion of instance features in traditional methods limit their scalability in terms of input persons. Therefore, we exploit a plug-and-play Instance Pooling module for multiple instance inputs (in Sec.~\ref{sec:ip-module}), which supports the recognition of group activities without surging computational costs.

\subsection{Expressive Keypoints representation}
\label{sec:expressive-keypoints}


\noindent \textbf{Data collection.} Benefiting from the dense landmarks provided by COCO-WholeBody~\cite{cocowholebody}, which encompasses $133$ keypoints, including $17$ keypoints for the body, $68$ for the face, $42$ for the hands, and $6$ for the feet, we have a base representation for fine-grained skeleton. In practice, COCO-WholeBody can be extracted from a top-down estimator. We firstly extract human bounding boxes using the ResNet50-based Faster-RCNN~\cite{fastrcnn}. Subsequently, the COCO-WholeBody~\cite{cocowholebody} keypoints within specified bounding boxes are obtained through the pre-trained human pose estimator~\cite{hrnet}.



\noindent \textbf{Keypoint selection.}
We observe directly using COCO-WholeBody as input not only incurred significant computational costs but also yielded lower performance, because there might be numerous redundant keypoints introducing substantial noise into the model. To alleviate this issue, we select the input 133 keypoints from two perspects. First, COCO-Wholebody not only includes body and detailed hand keypoints, but also includes face landmarks, which are intuitively not related to the human action. Besides, we analyze two statistical metrics: \textit{Video Variance} and \textit{Motion variance} on the NTU-120 dataset, which calculate the variance of keypoints for each person and motion frequency of each keypoint between frames, respectively. More details and results are provided in Sec.~\ref{supp:statistical}. We find facial keypoints (23-90th) have higher video variance and lower motion frequency, which indicates low contribution for action recognition. This observation guides us to manually remove them, resulting in the formation of the final Expressive Keypoints representation.

\subsection{Skeleton Transformation strategy}
\label{sec:gts}
The representation of Expressive Keypoints provides abundant motion cues for skeleton action recognition. However, directly feeding Expressive Keypoints into existing GCN methods encounters several limitations. \textbf{(i) Low efficiency:} Handling with much more limb joints significantly increases computational complexity compared to the coarse-grained ones. \textbf{(ii) Sub-optimal accuracy:} The topology graph of Expressive Keypoints is more complex and has multi-hop connections which hinders the network from effectively exchange information among distant nodes. Consequently, it faces a more pronounced long-range dependency problem~\cite{HDGCN}. We claim that the key problem is that \textit{traditional methods have a fixed skeleton structure during feed forward}.

To this end, we propose a novel Skeleton Transformation (SkeleT) strategy to gradually downsamples the Expressive Keypoints throughout the processing stages. The SkeleT strategy can be seamlessly integrated into most GCN methods to create our SkeleT-GCN (\textit{e.g} baseline: DGSTGCN~\cite{DGSTGCN} → ours: SkeleT-DGSTGCN) without modifying the inner implementation of their graph convolution and temporal convolution layers or the high-level architectural design. What we do is to encapsulating the baseline graph convolution layers within a proposed Grouped Mapping framework, where the input keypoint features are divided into groups and multiplied with the mapping matrices before being processed by the graph convolution layers. By strategically exploit Expressive Keypoints, our SkeleT-GCN can achieve comparable or even higher accuracy with much lower GFLOPs compared with its baseline GCN method.

\subsubsection{Preliminary and notations of GCN} 
The skeleton sequence $\mathbf{X} \in \mathbb{R}^{J\times T\times C}$ is defined by $J$ joints with $C$ dimension channels at each joint in $T$ frames. For most existing GCN-based methods, they share a same architecture design of $M$ spatial-temporal blocks, where each spatial-temporal block $\mathcal{F}$ contains a graph convolution layer $\mathcal{G}$ and a temporal convolution layer $\mathcal{T}$ to alternately model the spatial and temporal information. We use $\mathbb{B}=\{1,2,..,M\}$ to denote the index set of spatial-temporal blocks, which has two subset $\mathbb{B}^n$ and $\mathbb{B}^d$, where $\mathbb{B}^d$ contains the indices of downsampling blocks $\mathcal{F}^{d}$ that downsample the temporal length and $\mathbb{B}^n$ contains the indices of other normal blocks $\mathcal{F}^{n}$. The adjacent martix $\mathbf{A} \in \mathbb{R}^{J \times J}$ defines the topology links of human skeleton, where $\mathbf{A}_{ij}=1$ if $i$-th joint and $j$-th joint are physically connected and 0 otherwise. The computation of $\mathcal{F}$ can be summarized as:
\begin{equation}
\mathcal{F}(\mathbf{X},\mathbf{A}) = \mathcal{T}(\mathcal{G}(\mathbf{X},\mathbf{\widetilde{A}}))+\mathbf{X},
\end{equation}
\noindent where $\mathbf{\widetilde{A}} = \mathbf{A}+\mathbf{I}$ is the skeletal topology graph with added self-link.

\begin{figure}[t]
  \centering
  \includegraphics[width=0.9\linewidth]{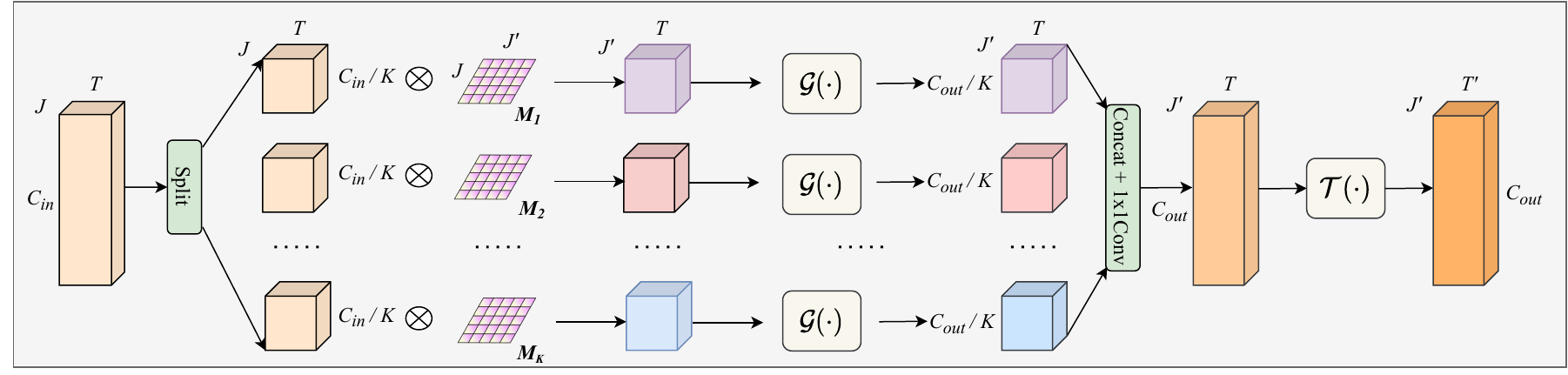}
  \caption{The architecture of Grouped Mapping Framework $\hat{\mathcal{F}}$. Most GCN-based methods' the graph convolution layer $\mathcal{G}$ and the temporal convolution layer $\mathcal{T}$ can be adopted.}
  \label{fig:gt-gc}
  \vspace{-5mm}
\end{figure}

\subsubsection{Grouped Mapping Framework}
To achieve the SkeleT strategy for existing GCN methods, we propose the Grouped Mapping Framework to encapsulate original graph convolution layers $\mathcal{G}$ and temporal convolution layers $\mathcal{T}$ of any GCN methods without modifying their inner design. The same high-level architecture $\mathbb{B}={\mathbb{B}}^n\cup{\mathbb{B}}^d$ is also inherited. We denote the Grouped Mapping Framework as $\hat{\mathcal{F}}$ and its detailed architecture is depicted in Fig.~\ref{fig:gt-gc}. Specifically, we split the channel dimension of the skeleton sequence $\mathbf{X}$ into $K$ groups, thereby reducing the channel width of each feature group to $C/K$.  Subsequently, each feature group is independently multiplied by a corresponding mapping matrix $\mathbf{M}$ to adaptively alter the skeleton structure. Next, we parallelize $K$ baseline graph convolution layers $\{ \mathcal{G}_1, ..., \mathcal{G}_K \}$ to extract group-specified features that can greatly enrich the motion feature representations across diverse structures. Finally, $K$ group features are concatenated along the channel dimension and processed by the baseline temporal convolution layer $\mathcal{T}$ to model the temporal dependency, generating the refined motion feature. The whole processing of our Grouped Mapping Framework $\hat{\mathcal{F}}$ can be formulated as follows:
\begin{equation}
\label{eq:gmf}
\hat{\mathcal{F}}(\mathbf{X},\mathbf{A},\mathbf{M}) = \mathcal{T}( \sigma( \mathcal{G}_k(\mathbf{M}_k\mathbf{X}_k,\mathbf{\widetilde{A}})\mathbf{W}) )+ res(\mathbf{X}), k \in\{1,...,K\},
\end{equation}
\noindent where $\mathbf{X}_k$ is the $k$-th split feature and $\mathbf{W}$ is a learnable weights. $\sigma(\cdot)$ and $res(\cdot)$ is the activation and residual connection, respectively. We provide further elaborations of mapping matrix $\mathbf{M}$ subsequently.

\textbf{Mapping matrix.} The main idea of downsampling the keypoints is achieved by being multiplied with the mapping matrix $\mathbf{M}^d \in\mathbb{R}^{J_{i} \times J_{i+1}}$ to fuse correlated joints. It maps the original skeleton $\mathbf{X}$ with $J_{i}$ joints to a new skeleton $\mathbf{X}^{\prime}$ with $J_{i+1}$ joints, which can be formulated as follows:
\begin{equation}
\mathbf{X}^{\prime}= \mathbf{M}^d \mathbf{X},
\end{equation}
Once the skeleton structure is downsampled, the new adjacent matrix can be calculated as follows:
\begin{equation}
\mathbf{A}^{\prime}=(\mathbf{M}^d)^T\mathbf{A}\mathbf{M}^d.
\end{equation}
The downsampling operation is only conduct in the downsampling blocks with indices in $\mathbb{B}^d$. For the other normal blocks in $\mathbb{B}^n$, the mapping matrix $\mathbf{M}^n \in\mathbb{R}^{J_{i} \times J_{i}}$ is defined as a learnable diagonal matrix that does not downsample the keypoints. It serves to re-weight the skeleton joints, enabling the network to prioritize important joints by allocating weights on the diagonal. Considering the index of $\mathcal{\hat{F}}$ and the type of mapping matrix, Eq.(\ref{eq:gmf}) can be detailed as follows: 
\begin{equation}
\hat{\mathcal{F}}_{(i)}(\mathbf{X},\mathbf{A},\mathbf{M})=\left\{
\begin{array}{lclc}
\mathcal{T}(\sigma( \{[\mathcal{G}_k(\mathbf{M}^n_k\mathbf{X}_k,\mathbf{\widetilde{A}})]\}_{k \in\{1,...,K\}}\mathbf{W})+\mathbf{X} & ,i \in \mathbb{B}^n,\\
\mathcal{T}(\sigma( \{[\mathcal{G}_k(\mathbf{M}^d_k\mathbf{X}_k,\mathbf{\widetilde{A}})]\}_{k \in\{1,...,K\}}\mathbf{W})+\mathbf{M}^d\mathbf{X} & , i \in \mathbb{B}^d.
\end{array}
\right.
\end{equation}

\begin{figure}[tbp]
  \centering
  \begin{minipage}{0.72\textwidth}
    \centering
    \includegraphics[width=\textwidth]{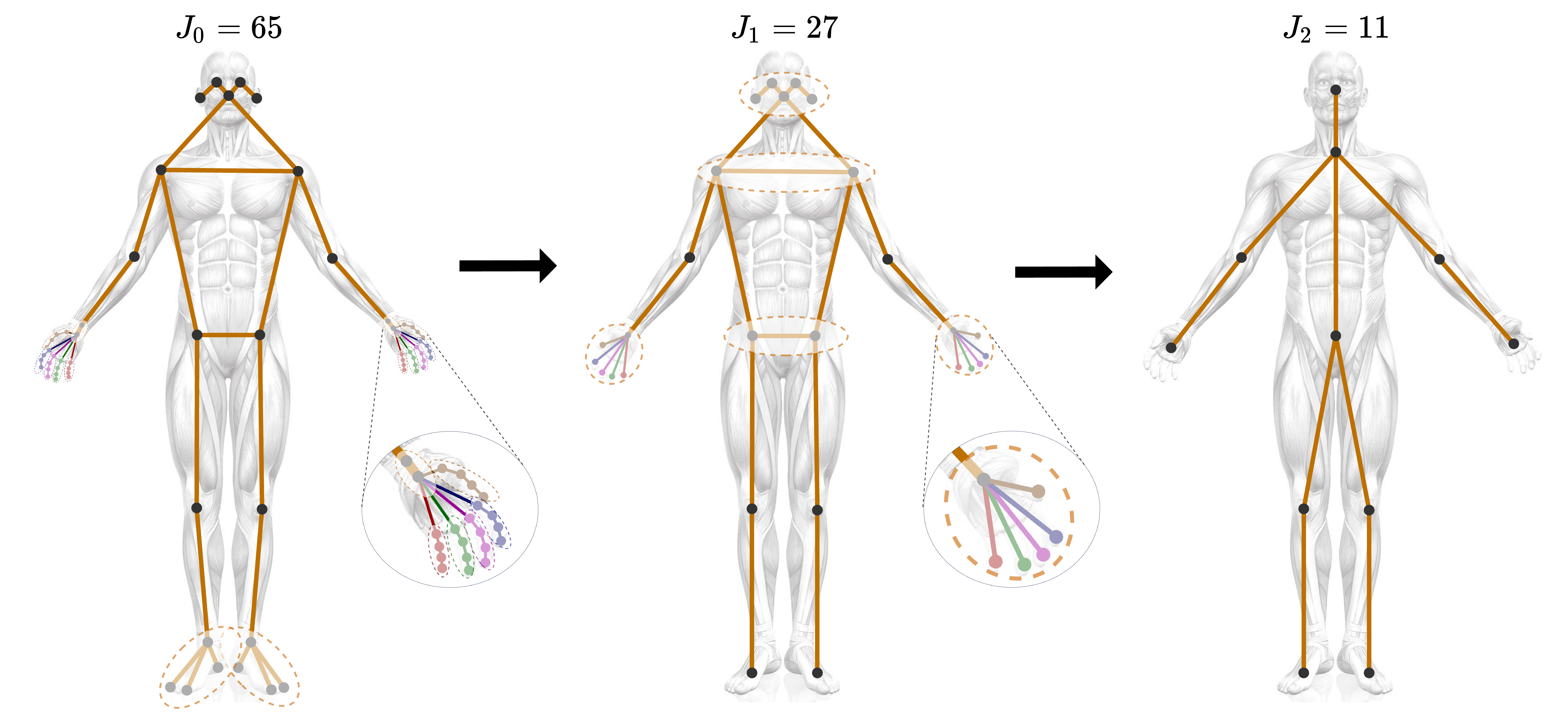}
    \caption{Pre-defined keypoint partition.}
    \label{fig:keypoint_partitioning}
  \end{minipage}
  \begin{minipage}{0.27\textwidth}
    \centering
    \includegraphics[width=\textwidth]{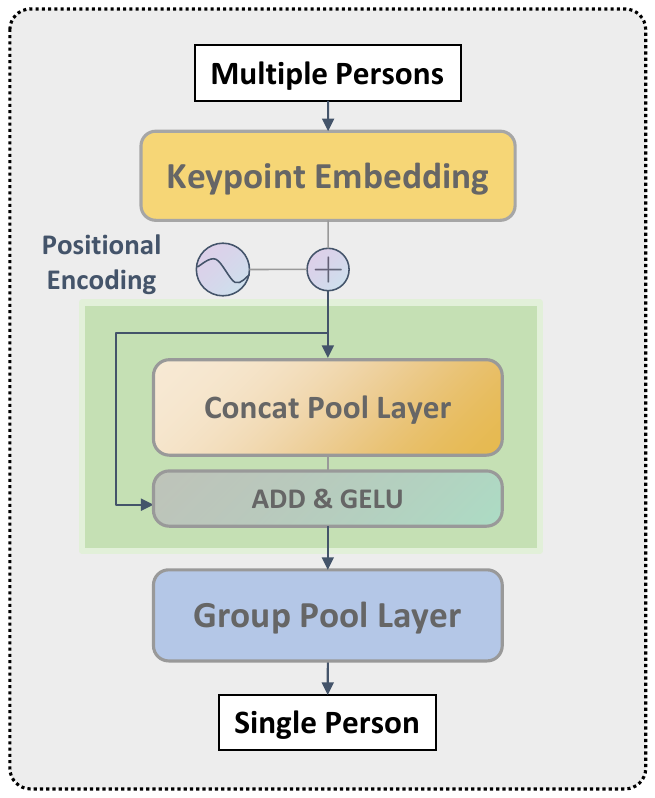}
    \caption{IP module.}
    \label{fig:ip}
  \end{minipage}
\end{figure}

\textbf{Pre-defined keypoint partition.}
As shown in Fig.~\ref{fig:keypoint_partitioning}, the above downsampling mapping matrix $\mathbf{M}^d$ has a weight of $[J_i, J_{i+1}]$ to map $J_i$ keypoints to $J_{i+1}$ keypoints, and it needs a good initialization to stabilize at beginning stage of training. 
Adjacent keypoints always have similar semantics for human action, therefore, we use the pre-defined semantical knowledge prior to initialize the $\mathbf{M}^d_{[i, i+1]}$. 
Specifically, $J_i$ joints can be divided into part set $\{P_{(i, i+1)}\}$, where $J_{i+1}^k$ ($k$-th joint of $J_{i+1}$) includes $P_{i, i+1}^k$ indexes of $J_i$ keypoints.
Once the partition is determined, the initialized element of $j$-th row, $k$-th column in $\mathbf{M}^d$ ($j \in J_{i}, k \in J_{i+1}$) can be formulated as follows:
\begin{equation}
\mathbf{M}^d_{(j, k)}=\{\begin{array}{cll}
\frac{1}{len(P_{(i, i+1)}^k)} & , j \in P_{i, i+1}^k, \\
0 & , \text { otherwise. }
\end{array}
\end{equation}
The keypoint partitions are semantically guided. Related joints like keypoints in the same finger are grouped as one part when initialization.


\subsection{Instance Pooling module}
\label{sec:ip-module}

The computation of previous GCN-based works scale linearly with the increasing number of persons in the video, making it less efficient for group activity recognition. The key problem is that traditional methods \textit{independently model each person's skeleton sequence} and then \textit{perform feature fusion at the late stage}.

To tackle this problem, we implement an plug-and-play Instance Pooling (IP) module which perform early feature fusion of the multiple input skeletons before feeding them to GCN. As illustrated in Fig.~\ref{fig:ip}, we obtain the keypoint embedding utilizing a fully connected layer and a keypoint positional encoding from the multi-person skeletal sequences. Subsequently, the Concat Pool Layer $\mathcal{P}_{c} (\cdot)$ and the Group Pool Layer $\mathcal{P}_{g} (\cdot)$ proposed by~\cite{skp} are adopted to aggregate $I$ instance-wise feature vectors. This process can be formulated as:
\begin{equation}
\mathbf{Y'}= \mathcal{P}_{g}(\sigma(\mathcal{P}_{c}(\mathbf{Y})+\mathbf{Y})),
\end{equation}
where $\mathbf{Y} = emb( \{{\mathbf{X}_1},\mathbf{X}_2,...,\mathbf{X}_I\}) \in \mathbb{R}^{I\times J\times T\times C}$ is the embedding of multi-person skeletons. $\mathbf{Y^{\prime}} \in \mathbb{R}^{J\times T\times C}$ is the aggregated single-person representation where the dimension of instance $I$ has been eliminated. Through early fusion in the lightweight IP module, the computationally burdensome spatial-temporal modeling will be conducted only once in the subsequent GCN, regardless of the number of input instances. The IP module serves as a flexible and lightweight extension for any GCN-based methods (including our SkeleT-GCN). It offers a a practical and efficient solution for extending GCN-based skeleton action recognition to multi-person group activity scenarios without surging computational cost.

\begin{table}[h]
  \centering
  \begin{minipage}{0.58\linewidth}
    \caption{Effectiveness of SkeleT strategy on Expressive Keypoints.}
    \centering
    \resizebox{\linewidth}{!}{
    \begin{tabular}{lclll}
    \toprule
    \multicolumn{1}{c}{Method} & \multicolumn{1}{c}{Format} & CS(\%) & CX(\%) & FLOPs\\
    \midrule
    STGCN++                      & NTU Keypoints         & 84.3 & 86.7  & 2.7G  \\
    STGCN++                      & Expressive Keypoints  & 92.6~\scriptsize\textcolor{red}{+8.3} & \textbf{94.5}~\scriptsize\textcolor{red}{+7.8}    & 6.9G  \\
    SkeleT-STGCN++                   & Expressive Keypoints  & \textbf{92.7} & \textbf{94.5}    & \textbf{2.6G}~\scriptsize\textcolor{blue}{-4.3}  \\ 
     \midrule 
    CTRGCN                       & NTU Keypoints         & 84.0 & 85.9    & 2.7G  \\
    CTRGCN                       & Expressive Keypoints  & \textbf{92.8}~\scriptsize\textcolor{red}{+8.8} & 94.5~\scriptsize\textcolor{red}{+8.6}   & 7.5G  \\
    SkeleT-CTRGCN                    & Expressive Keypoints  & \textbf{92.8} & \textbf{94.7}    & \textbf{2.5G}~\scriptsize\textcolor{blue}{-5.0} \\ 
    \midrule 
    DGSTGCN                      & NTU Keypoints         & 85.7 & 87.9    & \textbf{2.4G}  \\
    DGSTGCN                      & Expressive Keypoints  & 92.6~\scriptsize\textcolor{red}{+6.9} & 94.4~\scriptsize\textcolor{red}{+6.5}    & 6.3G  \\
    SkeleT-DGSTGCN                   & Expressive Keypoints  & \textbf{93.1} & \textbf{94.8}   & \textbf{2.4G}~\scriptsize\textcolor{blue}{-3.9}  \\ \bottomrule
    \end{tabular}}
    \label{tab:ablation-components}
  \end{minipage}
\begin{minipage}{0.41\linewidth}
    \centering
    \caption{Effectiveness of SkeleT on NTU.}
    \resizebox{\linewidth}{!}{
    \begin{tabular}{clccl}
    \toprule
    Format                      & \multicolumn{1}{c}{Method} & CS(\%) & CX(\%) & FLOPs \\ \midrule
    \multirow{6}{*}{{\begin{tabular}[c]{@{}c@{}}NTU\\      Keypoints\end{tabular}}}        
    & STGCN++                      & 84.3 & \textbf{86.7}      & 2.7G   \\
    & SkeleT-STGCN++                   & \textbf{84.9} & \textbf{86.7}      & 1.5G~\scriptsize\textcolor{blue}{-1.2}    \\ \cmidrule{2-5} 
    & CTRGCN                       & 84.0 & 85.9      & 2.7G    \\
    & SkeleT-CTRGCN                    & \textbf{84.1} & \textbf{86.4}      & 1.5G~\scriptsize\textcolor{blue}{-1.2}    \\ \cmidrule{2-5} 
    & DGSTGCN                      & \textbf{85.7} & \textbf{87.9}      & 2.4G    \\ 
    & SkeleT-DGSTGCN                   & \textbf{85.7} & 87.8      & 1.5G~\scriptsize\textcolor{blue}{-0.9}    \\ \bottomrule
    \end{tabular}}
    \label{tab:ablation-ntukpts}

    \caption{Effectiveness of IP module.}
    \resizebox{\linewidth}{!}{
    \begin{tabular}{rccc} \toprule
    \multicolumn{1}{c}{Method}        & Input persons & Accuracy(\%) & FLOPs \\ \midrule
    w/o IP module & 2             & 49.6         & 2.3G  \\
    w/~~ IP module  & 10            & \textbf{51.6}         & \textbf{1.5G}  \\ \bottomrule
    \end{tabular}}
    \label{tab:ablation-ip}
  \end{minipage}
  \hfill 
  \vspace{-2mm}
\end{table}

\section{Experiments}
We conduct comprehensive experiments to evaluate our proposed pipeline over seven datasets, including NTU-60~\cite{ntu60}, NTU-120~\cite{ntu120}, PKU-MMD~\cite{pku}, N-UCLA~\cite{ucla}, Kinetics-400~\cite{k400}, UCF-101~\cite{ucf101}, and HMDB-51~\cite{hmdb}. Overview of datasets (see Sec.~\ref{supp:datasets}) and implementation details (see Sec.~\ref{supp:implementation}) can be found in appendix. We report Top-1 accuracy to evaluate model's recognition performance, and report floating point operations (FLOPs) and number of parameters (Params.) to evaluate model's efficiency in terms of computation cost and model size.

\subsection{Effectiveness of proposed components}
\label{sec:ablation-effect}
We conduct evaluations for the effectiveness of every components in our proposed pipeline, which include the Expressive Keypoints representation, the SkeleT strategy, and the IP module. 



\begin{figure*}[h]
    \centering
    \includegraphics[width=0.95\textwidth]{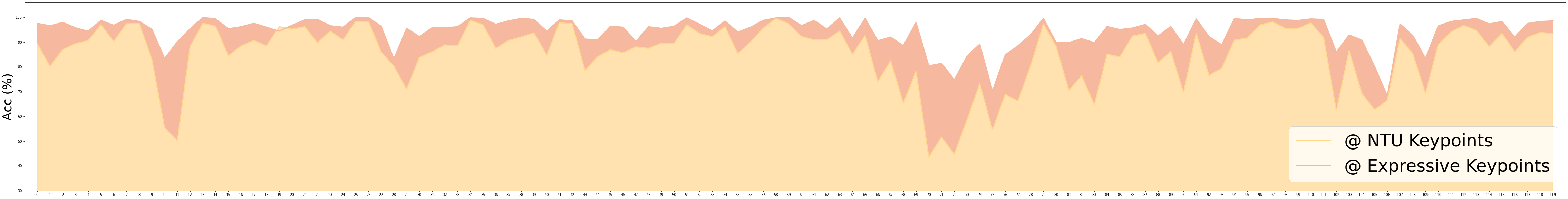}
    \caption{Accuracy comparison of 120 actions between NTU Keypoints and Expressive Keypoints.}
    \label{fig:ablation-wholebody1}
\end{figure*}

\begin{figure}[h]
  \centering
    \begin{minipage}{0.48\textwidth}
        \centering
        \includegraphics[width=\linewidth]{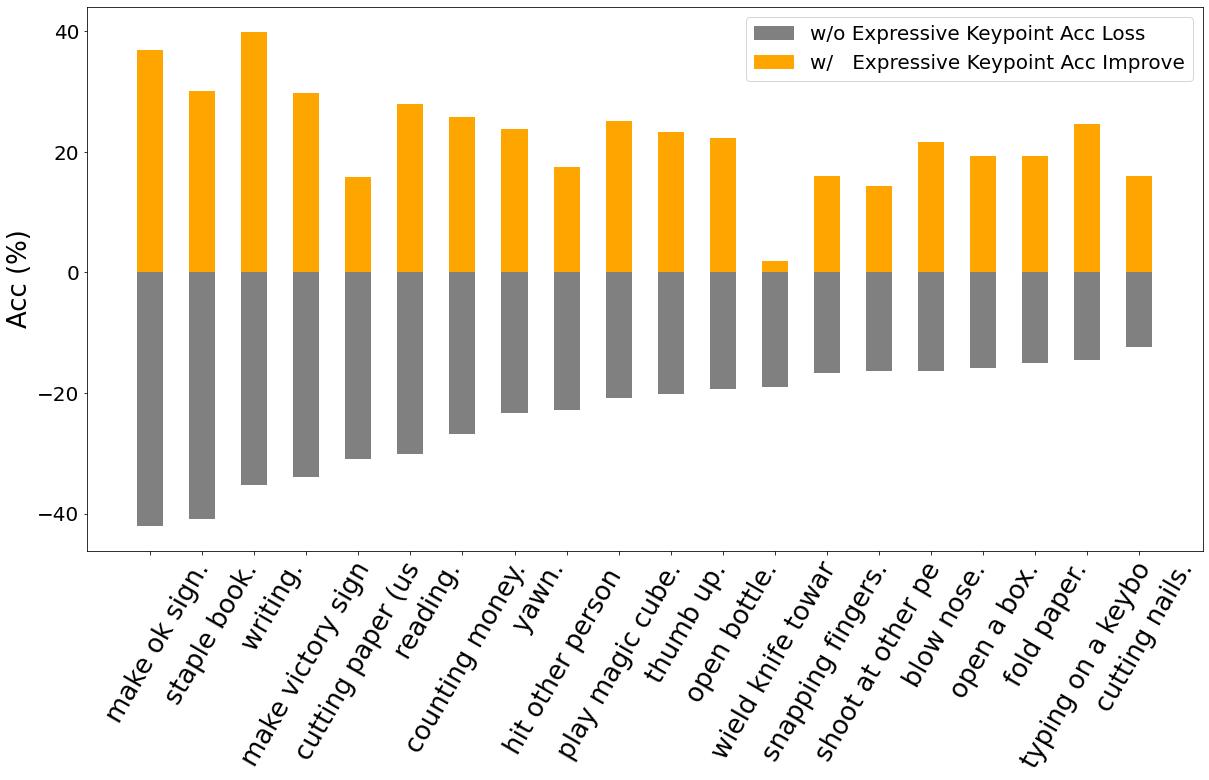}
        \caption{Comparison of top-20 hard cases. The \textcolor{gray}{gray} bar indicates reduced accuracy compared to average on NTU Keypoints, \textcolor{orange}{orange} bar denotes improved accuracy using Expressive Keypoints.}
        \label{fig:ablation-wholebody2}
    \end{minipage}
    \hfill
    \begin{minipage}{0.47\textwidth}
        \centering
        \includegraphics[width=\linewidth]{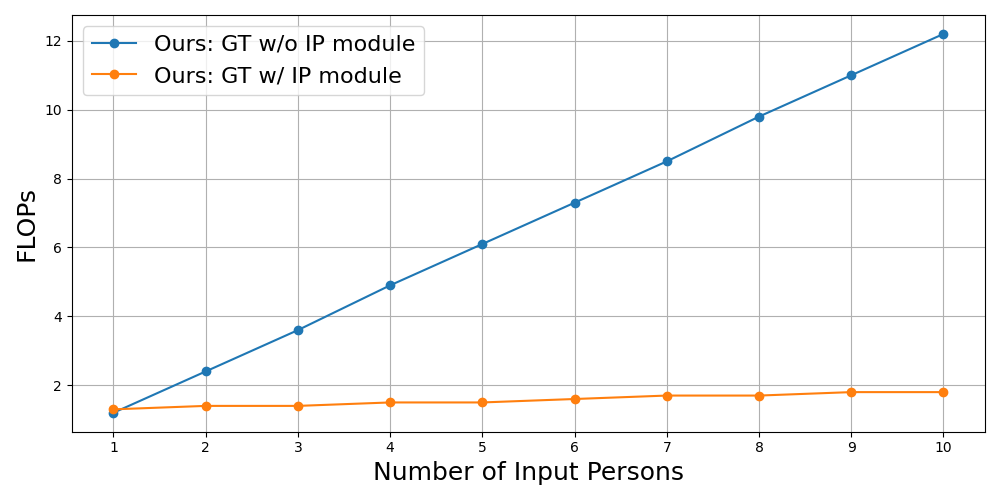}
        \caption{Ablation study on IP module with repect to the input person numbers. The FLOPs increases linearly with person number increasing without IP module. While the FLOPs hardly increases with IP module.}
        \label{fig:ablation-ip}
    \end{minipage}
    \vspace{-5mm}
\end{figure}

\textbf{Expressive Keypoints representation.}
On NTU-120, we directly feed Expressive Keypoints into three representative GCN methods, which are STGCN++~\cite{PYSKL}, CTRGCN~\cite{CTRGCN}, and DGSTGCN~\cite{DGSTGCN}. 
As shown in Tab.~\ref{tab:ablation-components}, the Expressive Keypoints representation significantly enhances action recognition performance on all three baseline networks (+7.8\%, +8.6\%, +6.5\%, respectively).
Additionally, we further assess the accuracy improvement on 120 action categories (Fig.~\ref{fig:ablation-wholebody1}) as well as the top-20 hard cases (Fig.~\ref{fig:ablation-wholebody2}) when replacing coarse-grained NTU keypoints with fine-grained Expressive Keypoints. It can be seen that incorporating detailed limb keypoints consistently boosts the skeleton action recognition performance especially for discerning those hard actions with nuanced limb movements.

\begin{table}[t]
  \vspace{-4mm}
  \centering
\begin{minipage}{0.42\linewidth}
    \centering
    \captionof{table}{Ablation study on input keypoint selection. \textit{Simple fingers} mean only one keypoint is retained for each finger.}
    \resizebox{\linewidth}{!}{
    \begin{tabular}{clccc}
    \toprule
    Protocol &~~~~Config of $V$                   & $N$         & Accuracy(\%)      & FLOPs         \\
    \midrule
    \#1 & COCO-WholeBody  & 133     & 93.4    & 12.8G \\
    \#2 & \#1+w/o face                     & 65        & \textbf{94.4}       & \textbf{6.3G}     \\ \midrule
    \#3 & \#2+w/o feet          & 59          & 94.1          & 5.8G          \\
    \#4 & \#2+simple fingers      & 35          & 90.6          & 3.4G          \\
    \#5 & \#2+w/o hands          & 23          & 88.0          & 2.4G          \\
    \#6 & \#2+\textbf{Ours: SkeleT} & 65 & \textbf{94.8} & \textbf{2.4G} \\
    \bottomrule
    \end{tabular}}
        \vspace{-2mm}
    \label{tab:ablation-keypoint}
\end{minipage}
  \hfill 
\begin{minipage}{0.23\linewidth}
    \centering
    \captionof{table}{Ablation study on the group configuration.}
    \resizebox{\linewidth}{!}{
    \begin{tabular}{ccccc}
    \toprule
    $K_0$ & $c$ & Config of $K$             & Acc(\%) \\ \midrule
    1 & 1 &{[}1, 1, 1{]} & 93.1   \\
    2 & 1 &{[}2, 2, 2{]} & 93.8   \\
    4 & 1 &{[}4, 4, 4{]} & 93.5   \\ \midrule
    1 & 2 &{[}1, 2, 4{]} & \textbf{94.8}   \\
    2 & 2 &{[}2, 4, 8{]} & 94.1   \\ 
    1 & 4 &{[}1, 4,16{]} & 93.9   \\ \bottomrule
    \end{tabular}}
    \label{tab:ablation-group}
    \vspace{-2mm}
\end{minipage}
\hfill 
  \begin{minipage}{0.295\linewidth}
  \centering
      \caption{Performance comparison on the PKU-MMD dataset.}
  \resizebox{\linewidth}{!}{
    \begin{tabular}{lc}
    \toprule
    Method        & PKU-MMD(\%) \\ \midrule
        Skeleton boxes~\cite{skl-boxes} & 54.8                       \\
    STA-LSTM~\cite{STA-LSTM}       & 86.9              \\
    HCN~\cite{HCN}            & 92.6            \\
    SRNet~\cite{SRNet}          & 93.1            \\ \midrule
    \textbf{Ours: SkeleT}*       & \textbf{98.4}       \\  \bottomrule
    \end{tabular}}
    \label{tab:pku}
    \vspace{-2mm}
  \end{minipage}
  \vspace{-4mm}
\end{table}

\textbf{SkeleT strategy.}
We further integrate proposed SkeleT strategy to the previous baseline GCN methods to form our SkeleT-GCN, which are SkeleT-STGCN++, SkeleT-CTRGCN, and SkeleT-DGSTGCN. By gradually downsampling Expressive Keypoints, three baseline models applying SkeleT strategy significantly reduce more than half of the computational cost (-4.3G, -5.0G, -3.9G) while achieving comparable or even higher accuracy, as shown in Tab.~\ref{tab:ablation-components}.
Moreover, we also evaluate the effectiveness of SkeleT strategy with NTU Keypoints input. As shown in Tab.~\ref{tab:ablation-ntukpts}, SkeleT strategy can also greatly reduce the computation cost~(from 2.4G$\sim$2.7G to 1.5G) of processing coarse-grained skeletal data while preserving accuracy. It can be observed that a slight accuracy drop occurs in one of the six settings. We consider this is because the coarse-grained skeletal representation is already very concise, and further downsampling might result in under-represented features.

\textbf{IP module.} On HMDB-51 which contains multi-person group activity scenarios, we use SkeleT-DGSTGCN to test the computational cost and accuracy with and without the IP module. The results are presented in Tab.~\ref{tab:ablation-ip}. We find that incorporating the IP module enhances recognition accuracy while considerably reducing the FLOPs. Moreover, Fig.~\ref{fig:ablation-ip} illustrates the variation in FLOPs with the number of input presons. Without the IP module, the computational cost escalates rapidly as the number of individuals increases due to the substantial feature modeling required for each individual in the traditional GCN pipeline. However, with the inclusion of the IP module, the increase in FLOPs is minimal since the features of multiple individuals are aggregated into a single representation by the lightweight IP module before fed into the subsequent GCN model.

\subsection{Configuration exploration}
\label{sec:ablation-config}

\textbf{Input keypoints selection.}
We extensively explore the selection of initial input keypoints. As shown in~Tab.~\ref{tab:ablation-keypoint}, experimental results demonstrate that removing facial keypoints from the COCO-WholeBody Keypoints (protocol \#1) to form our Expressive Keypoints (protocol \#2) is reasonable and aligns with the statistical analysis. Removing redundant points reduces the impact of introduced noise, resulting in higher accuracy with lower computational cost. Based on Expressive Keypoints, we try to further prune some keypoints. It is noticeable that removing the keypoints of limbs in a explicit way can achieve a decrease in FLOPs, but also incurs an equivalent drop in accuracy (protocol \#3$\sim$\#5). We argue that it is not applicable for explicitly selecting detailed limb keypoints in various actions of large-scale datasets. That is why we adopt a learning-based method SkeleT strategy for the implicit selection from Expressive Keypoints (protocol \#6), achieving great saving in FLOPs while maintaining high accuracy.

\textbf{Group design.}
~Tab.~\ref{tab:ablation-group} present six configurations in terms of the initial value of number of groups $K_0$ and group expand factor $c$. It is noticeable that static group designs ($c=1$) yield sub-optimal performances. For the expanding group designs, the [1, 2, 4] group configuration that can provide the best accuracy performance. We consider that too many groups will result in a small number of features after splitting the channels, limiting the representation ability.

\begin{table*}[t]
\caption{Accuracy and efficiency comparison with other SOTA methods on NTU-60 and NTU-120.}
\vspace{-1mm}
\centering
\resizebox{0.92\textwidth}{!}{
\begin{tabular}{@{}llcccccc@{}}
\toprule
\multicolumn{1}{l}{\multirow{2}{*}{Method}} & \multicolumn{1}{l}{\multirow{2}{*}{Modality}} & \multicolumn{2}{c}{NTU-60} & \multicolumn{2}{c}{NTU-120} & \multicolumn{2}{c}{Efficiency} \\ \cmidrule(l){3-8} 
\multicolumn{1}{c}{}                        & \multicolumn{1}{c}{}                          & CS(\%)       & CV(\%)      & CS(\%)       & CX(\%)       & FLOPs         & Params.        \\ \midrule
I3D~\cite{I3D}                                         & RGB                                           & —           & —          & 77.0         & 80.1         & 107.9G        & 12.1M          \\
GlimpseClouds~\cite{Glimpseclouds}                               & RGB                                           & 86.6         & 93.2        & —           & —           & 168.0G        & 46.8M          \\
VPN++~\cite{vpn++}                                       & RGB                                           & 91.9         & 94.9        & 86.7         & 89.3         & 112.1G        & 14.0M          \\
PoseC3D~\cite{PoseC3D}                                     & RGB                                           & 95.1         & —          & —           & —           & 41.8G         & 31.6M          \\ \midrule
P4Transformer~\cite{P4Transformer}                               & Point cloud                                   & 90.2         & 96.4        & 86.4         & 93.5         & 40.4G         & 44.1M          \\
PSTNet~\cite{PSTNet}                                      & Point cloud                                   & 90.5         & 96.5        & 87.0         & 93.8         & 54.1G         & 8.4M           \\
PST-Transformer~\cite{PST-Transformer}                             & Point cloud                                   & 91.0         & 96.4        & 87.5         & 94.0         & 48.8G         & 44.2M          \\ \midrule
PoseC3D~\cite{PoseC3D}                                     & Skeleton                                      & 94.1         & 96.9        & 86.9         & 90.3         & 20.9G         & 4.0M           \\
STGCN~\cite{STGCN}                                       & Skeleton                                      & 90.7         & 96.5        & 86.2         & 88.4         & 21.4G         & 12.3M          \\
AAGCN~\cite{MSAAGCN}                                       & Skeleton                                      & 91.5         & 96.7        & 86.9         & 88.8         & 24.3G         & 15.1M          \\
MSG3D~\cite{MSG3D}                                       & Skeleton                                      & 91.7         & 96.9        & 87.9         & 89.6         & 41.1G         & 12.7M          \\
STGCN++~\cite{PYSKL}                                     & Skeleton                                      & 92.1         & 97.0        & 87.5         & 89.8         & 10.6G         & 5.5M           \\
CTRGCN~\cite{CTRGCN}                                      & Skeleton                                      & 92.1         & 97.0        & 88.1         & 89.9         & 10.8G         & 5.6M           \\
DGSTGCN~\cite{DGSTGCN}                                     & Skeleton                                      & 93.2         & 97.5        & 89.6         & 91.4         & 9.6G          & 6.6M           \\
ShiftGCN~\cite{ShiftGCN}                                    & Skeleton                                      & 90.7         & 96.5        & 85.9         & 87.6         & 10.0G         & 2.8M           \\
InfoGCN~\cite{InfoGCN}                                     & Skeleton                                      & 93.0         & 97.1        & 89.8         & 91.2         & 10.0G         & 9.4M           \\
HDGCN~\cite{HDGCN}                                       & Skeleton                                      & 93.4         & 97.2        & 90.1         & 91.6         & 9.6G          & 10.1M          \\ \midrule
\textbf{Ours: Baseline}*                              & Skeleton(+limb details)                       & 96.9         & \textbf{99.6}        & 94.3         & 96.1         & 25.0G         & 6.6M           \\
\textbf{Ours: SkeleT}*                                & Skeleton(+limb details)                       & \textbf{97.0}         & \textbf{99.6}        & \textbf{94.6}         & \textbf{96.4}         & 9.6G          & 5.2M           \\ \bottomrule
\end{tabular}
}
\label{tab:ntu}
\end{table*}

\subsection{Comparison with the state-of-the-art}

When comparing to the state-of-the-art (SOTA), we choose DGSTGCN~\cite{DGSTGCN} with Expressive Keypoints input as the baseline method (denoted as \textbf{Ours: Baseline}), and apply SkeleT strategy to form our SkeleT-DGSTGCN (denoted as \textbf{Ours: SkeleT}). In experiments, * indicates using Expressive Keypoints, we adopted a 4-stream fusion strategy similar to the previous works \cite{MSAAGCN,DGSTGCN,CTRGCN,PYSKL}.

On NTU-60 and NTU-120, as shown in Tab.~\ref{tab:ntu}, Expressive Keypoints greatly improves the accuracy for skeleton-based action recognition, even surpassing the SOTA point cloud-based~\cite{PST-Transformer} and RGB-based methods~\cite{PoseC3D}. Upon applying the SkeleT strategy, our method achieves significant savings in the computational cost (25.0G $\rightarrow$ 9.6G), with comparable or even higher accuracy.


On PKU-MMD, Tab.~\ref{tab:pku} shows our method outperforming all the previous keleton-based methods by a noticeable margin, achieving the state-of-the-art performance with the top-1 accuracy of $98.4\%$.

On N-UCLA, as showed in Tab.~\ref{tab:ucla}, our method achieves $97.6\%$ top-1 accuracy, which also surpasses the previous best method~\cite{HDGCN}. It is notable that, among the standard skeleton-based datasets, N-UCLA has the most significant variations in viewpoint and severe occlusions. Despite being limited by the estimated 2D representation that is unable to leverage depth information and 3D spatial augmentations (\textit{e.g.} 3D random rotation), our approach still reaches a very promising performance.

We further extending SkeleT-DGSTGCN with the IP module (denoted as \textbf{Ours: SkeleT+IP}), which allows for evaluating our method on the more general in-the-wild action recognition datasets~\cite{k400,ucf101,hmdb}. For Kinetic-400 that encompass many human-object interaction scenarios, such as \textit{peeling apples} and \textit{peeling potatoes}, the accuracy of pure skeleton-based methods on the Kinetic-400 is far below than other datasets since they lack of capturing object information. As a result, SKP~\cite{skp} resorts to incorporating object contours and improves the accuracy of keypoint-based benchmark to $52.3\%$. However, as showed in Tab.~\ref{tab:k400}, by strategically utilizing Expressive Keypoints, our method achieves the SOTA performance ($53.1\%$) on the Kinetics-400 dataset even without the object information. This is made possible through our expressive skeletal representation and effective transformation strategy, demonstrating the effectiveness of our pipeline even under these challenging conditions.

Moreover, we provide an apple-to-apple comparison on UCF-101 and HMDB-51. As demonstrated in Tab.~\ref{tab:ucf-hmdb}, our method consistently surpasses the previous skeleton-based SOTA methods~\cite{PoseC3D,skp} regardless of whether pre-training is conducted on the Kinetics-400 dataset or not.

\begin{table}[t]
  \vspace{-4mm}
  \centering
  \begin{minipage}{0.295\linewidth}
  \centering
    \caption{Performance comparison on N-UCLA.}
  \resizebox{\linewidth}{!}{
    \begin{tabular}{lc}
    \toprule
    Method~~~~~~~~~~~~~~~           & N-UCLA(\%) \\ \midrule
    CTRGCN~\cite{CTRGCN}             & 96.5         \\
    InfoGCN~\cite{InfoGCN}            & 97.0         \\
    HDGCN~\cite{HDGCN}              & 97.2         \\ \midrule
    \textbf{Ours: SkeleT}*         & \textbf{97.6}         \\ \bottomrule
    \end{tabular}}
    \vspace{-2mm}
    \label{tab:ucla}
  \end{minipage}
  \hfill 
\begin{minipage}{0.31\linewidth}
    \centering
    \caption{Performance comparison on Kinetics-400.}
    \resizebox{\linewidth}{!}{
    \begin{tabular}{lc}
    \toprule
    Method         & Kinetics-400(\%) \\ \midrule
    STGCN~\cite{STGCN}                             & 30.7             \\
    MSG3D~\cite{MSG3D}                             & 38.0             \\
    PoseC3D~\cite{PoseC3D}                            & 47.7             \\
    SKP~\cite{skp} (w/ objects)                   & 52.3             \\ \midrule 
    \textbf{Ours: SkeleT+IP}* & \textbf{53.1}   \\ \bottomrule
    \end{tabular}}
    \vspace{-2mm}
    \label{tab:k400}
\end{minipage}
  \centering
 \begin{minipage}{0.365\linewidth}
    \centering
    \caption{Performance comparison on UCF-101 and HMDB-51.}
    \resizebox{\linewidth}{!}{
    \begin{tabular}{lccc}
    \toprule
    \multirow{2}{*}{Method}   &   Kinetics-400~      & UCF-101   & HMDB-51   \\ 
    & Pretraining~ & (\%) & (\%) \\ \midrule
    Potion~\cite{potion}     &  \ding{53}                       & 65.2          & 43.7          \\
    PoseC3D~\cite{PoseC3D}  &  \ding{53}                          & 79.1          & 58.6          \\ 
    \textbf{Ours: SkeleT+IP}*   &  \ding{53}   & \textbf{82.5} & \textbf{60.1} \\ \midrule
    PoseC3D~\cite{PoseC3D}     &  
    \ding{51}                      & 87.0          & 69.3          \\
    SKP~\cite{skp} (w/ objects)     &  
    \ding{51}             & 87.8          & 70.9          \\ 
    \textbf{Ours: SkeleT+IP}* &  
    \ding{51} & \textbf{88.7} & \textbf{74.6} \\ \bottomrule
    \end{tabular}}
    \vspace{-2mm}
    \label{tab:ucf-hmdb}
  \end{minipage}
  \vspace{-4mm}
\end{table}

\section{Conclusion}
\label{sec:conclusion}
In this work, we propose the Skeleton Transformation strategy using the Expressive Keypoints representation to achieve high performance in discriminating detailed actions while maintaining the high efficiency. Furthermore, we implement an Instance Pooling module, expanding the applicability of GCN-based methods to multi-person scenarios. Comprehensive experiments over seven datasets demonstrate our pipeline's superior performance and robust generalization.

\newpage
{
    \small
    \bibliographystyle{ieeenat_fullname}
    \bibliography{neurips_2024}
}

\newpage
\appendix
\section*{Appendix}
In this appendix, we provide overview and visualization of datasets, implementation details, additional experimental results, limitations and broader impact of our method to complement the main paper.

\section{Overview of datasets}
\label{supp:datasets}
We conduct comprehensive experiments to evaluate our proposed pipeline over seven datasets, which are NTU-60~\cite{ntu60}, NTU-120~\cite{ntu120}, PKU-MMD~\cite{pku}, N-UCLA~\cite{ucla}, Kinetics-400~\cite{k400}, UCF-101~\cite{ucf101}, and HMDB-51~\cite{hmdb}. 

\textbf{NTU-60} and \textbf{NTU-120} can be can be collectively referred to as \textbf{NTU RGB+D}, which is currently the largest dataset for skeleton human action recognition. The NTU-60 dataset contains 56,880 videos of 60 human actions. The authors of this dataset recommend two split protocols: CS and cross-view (CV). The NTU-120 dataset is a superset of NTU-60 and contains a total of 113,945 samples over 120 classes. The authors of this dataset recommend two split protocols: cross-subject (CS) and cross-set (CX). We conduct experiments on NTU-60 and NTU-120 following those recommended protocols.

\textbf{PKU-MMD} dataset is originally proposed for action detection. For the action recognition task, we crop long videos to get short clips based on the temporal annotations following~\cite{ISC}. The PKU-MMD has nearly 20,000 action instances over 51 classes.

We follow the recommended CS split protocol for training and testing.

\vspace{5mm}
\begin{figure*}[h]
    \centering
    \resizebox{1.0\linewidth}{!}{
    \includegraphics{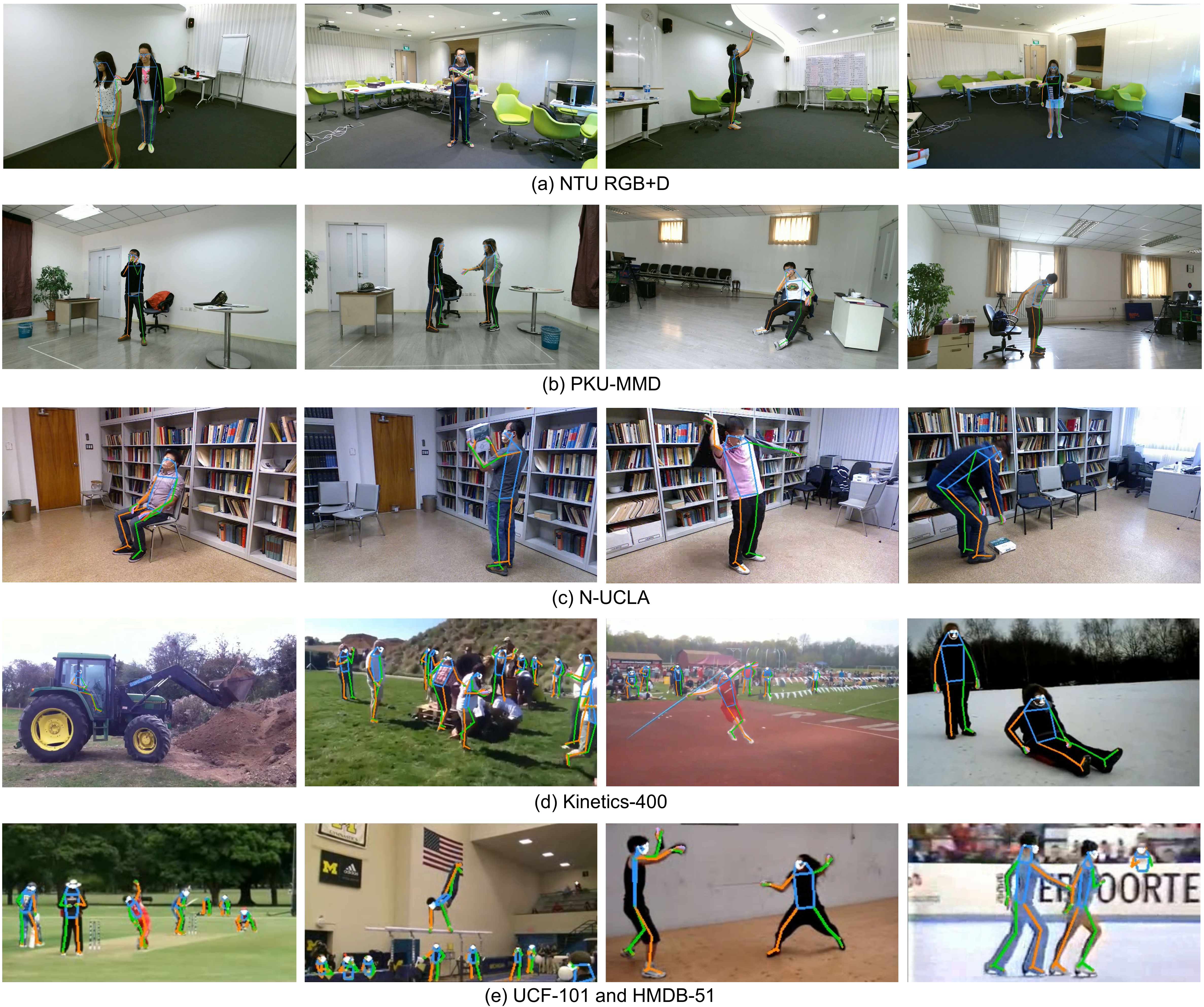}}
    \caption{Visualization of extracted whole-body poses from datasets~\cite{ntu60,ntu120,pku,ucla,k400,ucf101,hmdb}}
    \label{fig:dataset-vis}
\end{figure*}
\vspace{5mm}

\textbf{N-UCLA} contains 1494 video clips covering 10 action categories, which are performed by 10 different subjects. It has the most various significant variations in viewpoint and severe occlusions. We follow the same evaluation protocol in~\cite{CTRGCN}.

\textbf{Kinetics-400}, \textbf{UCF-101}, and \textbf{HMDB-51} are general action recognition datasets collect from web. With the incorporation of the Instance Pooling module, we have extended our pipeline to these in-the-wild datasets. The Kinetics-400 is a large-scale video dataset with 300,000 videos and 400 action classes. The UCF-101 dataset comprises approximately 13,000 videos sourced from YouTube, categorized into 101 action labels. The HMDB-51 consists of around 6,700 videos with 51 actions.


\section{Visualization of the extracted whole-body poses}
\label{supp:visualization}
We visualize the extracted poses of the aforementioned seven datasets~\cite{ntu60,ntu120,pku,ucla,k400,ucf101,hmdb}.

\textbf{NTU RGB+D} and \textbf{PKU-MMD} datasets are notable for (1) high resolution and excellent image quality and (2) containing at most two people, free from interference by individuals unrelated to the task. Consequently, the quality of the estimated poses is very high, as shown in Fig.~\ref{fig:dataset-vis}\textcolor{red}{a} and Fig.~\ref{fig:dataset-vis}\textcolor{red}{b}.

\textbf{N-UCLA} dataset is also shot indoors, the image quality is relatively high, resulting in fairly good quality pose estimations depicted in~Fig.~\ref{fig:dataset-vis}\textcolor{red}{c}. In contrast to NTU RGB+D and PKU-MMD, N-UCLA does not have dual-person actions and focuses solely on single-person action recognition.

\begin{figure}[t]
\vspace{4mm}
    \centering
    \resizebox{0.99\linewidth}{!}{
        \includegraphics{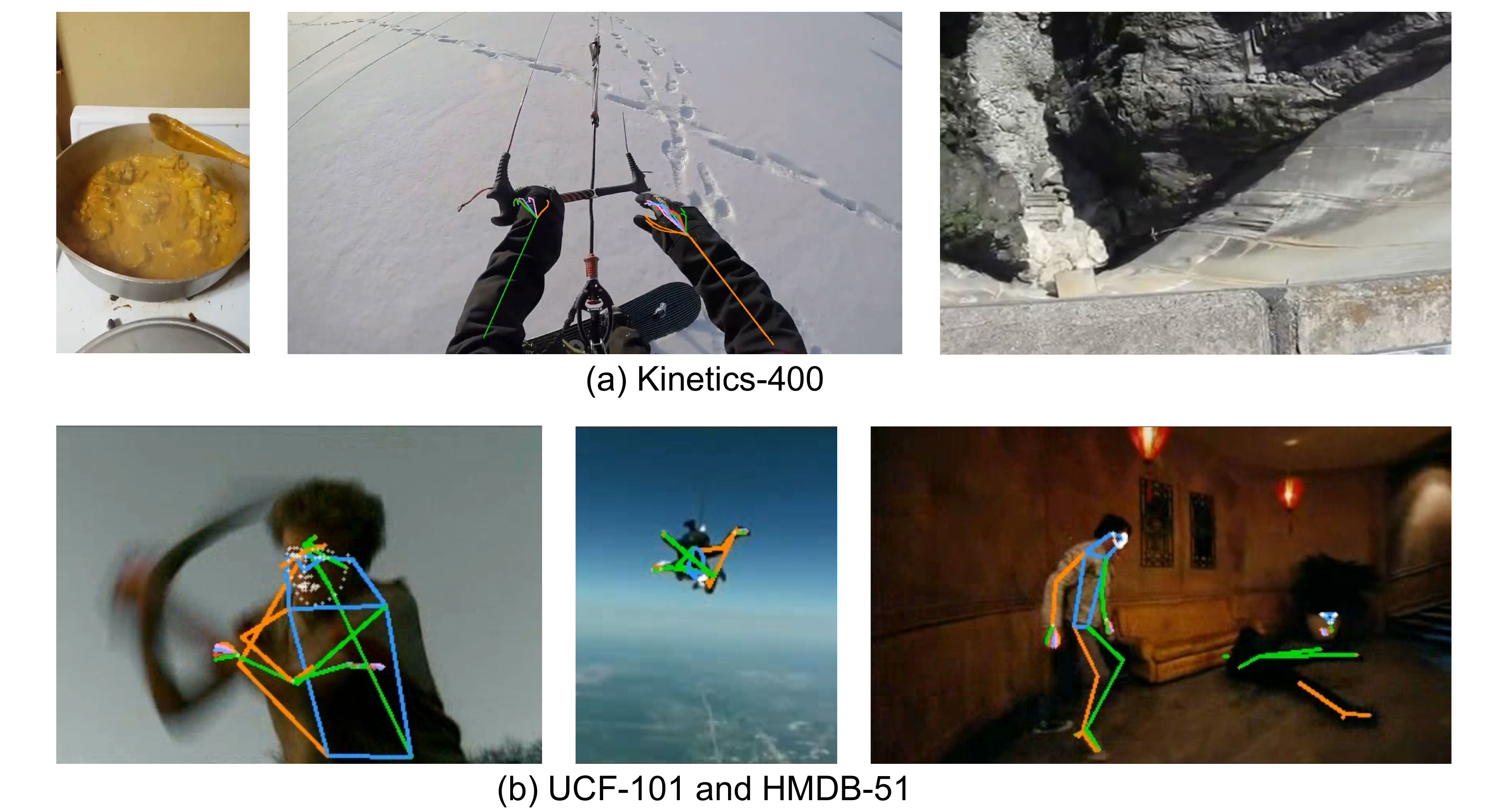}}
\caption{\textbf{(a).} Visualization of the poor-quality pose estimation results on the Kinetics-400 dataset. Left: Human missing during the action \textit{cooking chicken}. Middle: Only part of the human body appears during the action \textit{snowkiting}. Right: Human skeleton is too small to be recognized in the action \textit{bungee jumping}. \textbf{(b).} Visualization of the poor-quality pose estimation results on the UCF-101 and HMDB-51 datasets.}
    \label{fig:bad-vis}
    \vspace{2mm}
\end{figure}

\textbf{Kinetics-400} is a large-scale in-the-wild video action recognition dataset presenting complex scenes with numerous multi-person actions (crowd actions) and frequent appearances of unrelated individuals. We provide some examples that our estimator accurately predicts the human poses in~Fig.~\ref{fig:dataset-vis}\textcolor{red}{d}. However, since it is not human-centric, there are some problems that will degrade the quality of the extracted skeleton, as shown visualized in~Fig.~\ref{fig:bad-vis}\textcolor{red}{a}.

\textbf{UCF-101} and \textbf{HMDB-51} datasets are also in-the-wild video action recognition datasets, where the locations, scales, and number of persons may vary a lot. Fig.~\ref{fig:dataset-vis}\textcolor{red}{e} demonstrates some extracted poses with relatively good quality. However, due to low video resolution, tiny persons, and significant motion blur, the quality of most extracted poses is quite low, as shown in~Fig.~\ref{fig:bad-vis}\textcolor{red}{b}.

\newpage
\begin{figure*}[t]
    \centering
    \begin{subfigure}{0.96\textwidth}
        \centering
        \includegraphics[width=\textwidth]{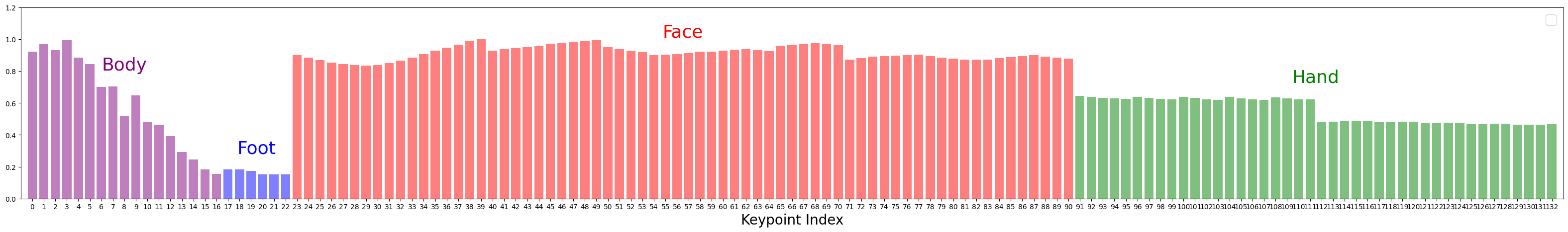}
        \caption{Video variance distribution}
        \label{fig:statistic-sub1}
    \end{subfigure}
    
    \begin{subfigure}{0.96\textwidth}
        \centering
        \includegraphics[width=\textwidth]{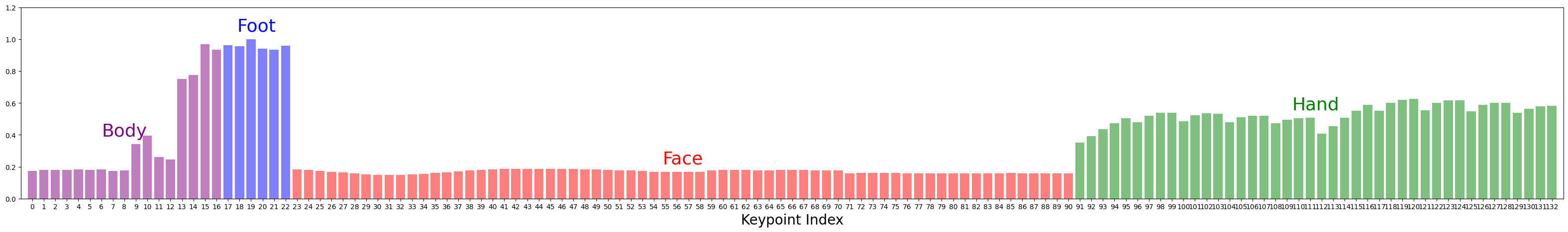}
        \caption{Motion variance distribution}
        \label{fig:statistic-sub2}
    \end{subfigure}
    
    \caption{Statistical results of whole-body keypoints on the NTU-120 dataset.}
    \label{fig:statistic}
    \vspace{-2mm}
\end{figure*}
\section{Statistical metrics and results}
\label{supp:statistical}
We conduct statically analysis on NTU-120 dataset, which involves two specific statistical metrics: \textbf{(i)} \textit{Video Variance $Var^{v}_{i}$}, calculates the variance of keypoints for each person across all videos. A lower value of $Var^{v}_{i}$ is indicative of a keypoint distribution that is more consistent and, consequently, more amenable:
\begin{equation}
    \label{eq:instance_variance}
    Var^{v}_{i} = \frac{1}{S} \sum_{s=1}^{S}(v_{i,s} - \overline{\mu}_{vi})^2,
\end{equation}
where $S$ represents number of videos, $v_{i,s}$ is mean of $i$-th joint positions in each video $s$, and $\overline{\mu}_{vi}$ indicates mean of all $v_{i,s}$.

\textbf{(ii)} \textit{Motion variance $Var^{m}_{i}$}, measures the motion frequency and range of each keypoint between frames, where higher $Var^{m}_{i}$ indicates more obvious movement for action recognition. 
\begin{equation}
    \label{eq:motion_variance}
    Var^{m}_{i} = f_{\sigma}(\frac{1}{T-1} \sum_{t=1}^{T-1} \frac{\sqrt{(p_{i,t+1} - p_{i,t})^2}}{\epsilon_i}),
\end{equation}
where $f_{\sigma}$ denotes the standard deviation function computed across videos, $p_{i,t+1}$ indicates $i$-th keypoint position in the $t$-th frame, and $\epsilon_i$ is area scale coefficient of different parts, which is used to normalize the motion variance. 

As illustrated in~Fig.\ref{fig:statistic}, facial keypoints (23-90th) have higher video variance and lower motion frequency, which indicates low contribution for action recognition. This observation guides us to manually remove them.

\section{Overall architecture of our SkeleT-GCN}
\label{supp:statistical}
Three representative GCN methods are adopted to be our baseline model, which are STGCN++, CTRGCN, DGSTGCN. All these models share the same high-level design. We apply the SkeleT strategy to form our corresponding SkeleT-GCN, which are SkeleT-STGCN++, SkeleT-CTRGCN, and SkeleT-DGSTGCN, respectively. 

The integration of the SkeleT strategy is seamless, thus the same overall architecture is inherited. Which includes 10 spatial-temporal blocks, and the output channels (number of features) for each block are configured as 64, 64, 64, 64, 128, 128, 128, 256, 256, and 256, respectively. The 5th and 8th blocks are downsampling blocks, while the other blocks are normal blocks. In each downsampling block, the groups expand at a factor of 2, the temporal length is reduced to half, and the number of joints is downsampled from 65 to 27 and futher to 11. Through a 2D Avg-Pooling, the temporal and joint dimensions are eliminated and the output is used by the classifier to predict a score vector for video-level action recognition.
\newpage

\section{Implementation details}
\label{supp:implementation}
\subsection{Hyperparameters}
Following the good practices of PYSKL~\cite{PYSKL}, we use the same hyperparameter setting for all GCN models to ensure fair comparison. Specifically, we employ the Stochastic Gradient Descent with a Nesterov momentum of 0.9 and weight decay of 0.0005. When training from scratch, the initial learning rate is set to 0.1, and we train all models for 120 epochs with the Cosine Annealing LR scheduler. On the UCF-101 and HMDB-51 datasets, we fine-tune all models based on the Kinetics-400 pretrained weights for 120 epochs with a initial learning rate of 0.01, which will decay with a factor 0.1 at epoch 90 and 110. The hyperparameters of batch size, temporal length, and number of input persons employed for each datasets are listed in~Tab.~\ref{tab:dataset}. We use zero-padding or cropping for each video to satisfy the fixed number of input persons. Our models are implemented with the PyTorch deep learning framework. All the experiments are conduct on a single Linux server with four RTX 3090 GPUs for distributed training and testing.

\begin{table*}[t]
\centering
\caption{Hyperparameters and augmentation of each dataset during training.}
\vspace{-2mm}
\resizebox{0.98\linewidth}{!}{
\begin{tabular}{c|cccccc}
\toprule
                    & ~NTU-RGB+D~ & PKU-MMD & N-UCLA            & Kinetics-400 & UCF-101 & HMDB-51           \\ \midrule
Optimizer           & \multicolumn{6}{c}{Stochastic Gradient Descent}                                       \\
Number of epochs    & \multicolumn{6}{c}{120}                                                              \\
Number of persons          & 2       & 2      & 1                & 10          & 10      & 10                \\
Temporal length          & 100       & 100      & 50                & 100          & 100      & 100                \\
Batch size          & 128       & 64      & 16                & 128          & 64      & 64                \\
Pretraining dataset & ~~~~~~None~~~~~~      & ~~~~~~None~~~~~~    & ~~~~~~None~~~~~~         & ~~~~~~None~~~~~~         & ~~~Kinetics-400~~~    & ~Kinetics-400~      \\ 
Learning rate       & \multicolumn{4}{c}{\cellcolor{gray!10}{0.1}}                & \multicolumn{2}{c}{\cellcolor{gray!25}{0.01}}            \\
Scheduler           & \multicolumn{4}{c}{\cellcolor{gray!10}{Cosine Annealing}}                & \multicolumn{2}{c}{\cellcolor{gray!25}{Step [90, 110]}}        \\
Weight decay        & \multicolumn{6}{c}{0.0005}                                                           \\
Momentum            & \multicolumn{6}{c}{Nesterov, 0.9}                                                    \\ 
\midrule
Random scaling      & \multicolumn{3}{c}{\cellcolor{gray!10}{None}}                & \multicolumn{3}{c}{\cellcolor{gray!25}{{[}0.85, 1.15{]}}}       \\
Random cropping     & \multicolumn{3}{c}{\cellcolor{gray!10}{None}}                & \multicolumn{3}{c}{\cellcolor{gray!25}{{[}0.56, 1.00{]}}}       \\
Random flipping     & \multicolumn{3}{c}{\cellcolor{gray!10}{None}}                & \multicolumn{3}{c}{\cellcolor{gray!25}{0.5}}                    \\
Temporal sampling   & \multicolumn{6}{c}{Uniform Sampling}                                                 \\ \bottomrule
\end{tabular}
}
\label{tab:dataset}
\end{table*}

\subsection{Data augmentation}
Uniform Sampling~\cite{PoseC3D} is adopted as a strong temporal augmentation strategy, which evenly partitions the original skeleton sequence into $T$ splits and randomly extracts one frame from each split to form a clip of length $T$.
On the NTU RGB+D, PKU-MMD, and N-UCLA datasets, no spatial augmentation is utilized for processing 2D Expressive Keypoints.
On the Kinetics-400, UCF-101, and HMDB-51 datasets, we employ substantial spatial data augmentations, \textit{e.g.} random scaling, cropping, and flipping the keypoints. Detailed augmentation for each datasets are listed in~Tab.~\ref{tab:dataset}.

\begin{table*}[b]
\centering
\vspace{-1mm}
\caption{We benchmark GCN skeleton-based action recognition methods on the NTU-60 dataset. The weights assigned to components of 2s-fusion and 4s-fusion are [1:1] and [3:3:2:2], respectively.}
\vspace{-1mm}
\resizebox{0.98\linewidth}{!}{
\begin{tabular}{l|cccc|cccc|cc}
\toprule
\multicolumn{1}{l|}{\multirow{2}{*}{Method}} & \multicolumn{4}{c|}{NTU-60 CS}                              & \multicolumn{4}{c|}{NTU-60 CV}                           & \multicolumn{2}{c}{Efficiency} \\ \cmidrule{2-11} 
\multicolumn{1}{c|}{}                        & Joint(\%)         & Bone(\%)          & ~~2s(\%)~~            & ~~4s(\%)~~            & Joint(\%)         & Bone(\%)          & ~~2s(\%)~~            & ~~4s(\%)~~            & FLOPs          & Params        \\ \midrule
STGCN++                                      & 95.6          & 95.8          & 96.5          & 96.8          & \textbf{99.1} & 98.9          & \textbf{99.4} & \textbf{99.5} & 6.9G           & 1.4M          \\
SkeleT-STGCN++                                   & \textbf{95.7} & \textbf{95.9} & \textbf{96.6} & \textbf{97.0} & \textbf{99.1} & \textbf{99.0} & \textbf{99.4} & \textbf{99.5} & \textbf{2.6G}  & \textbf{1.2M} \\ \midrule
CTRGCN                                       & 95.8          & \textbf{96.2} & 96.7          & 96.9          & 99.0          & \textbf{99.0} & 99.4          & \textbf{99.5} & 7.5G           & 1.4M          \\
SkeleT-CTRGCN                                    & \textbf{96.0} & \textbf{96.2} & \textbf{97.0} & \textbf{97.1} & \textbf{99.2} & \textbf{99.0} & \textbf{99.5} & \textbf{99.5} & \textbf{2.5G}  & \textbf{1.1M} \\ \midrule
DGSTGCN                                      & 95.1          & 95.8          & 96.6          & 96.9          & \textbf{99.3} & \textbf{99.1} & \textbf{99.5} & \textbf{99.6} & 6.3G           & 1.6M          \\
SkeleT-DGSTGCN                                   & \textbf{95.8} & \textbf{96.0} & \textbf{96.7} & \textbf{97.0} & \textbf{99.3} & \textbf{99.1} & \textbf{99.5} & \textbf{99.6} & \textbf{2.4G}  & \textbf{1.3M} \\ \bottomrule
\end{tabular}
}
\label{tab:benchmark-ntu}
\end{table*}

\begin{table*}[t]
\centering
\vspace{-1mm}
\caption{We benchmark GCN skeleton-based action recognition methods on the NTU-120 dataset. The weights assigned to components of 2s-fusion and 4s-fusion are [1:1] and [3:3:2:2], respectively.}
\vspace{-1mm}
\resizebox{0.98\linewidth}{!}{
\begin{tabular}{l|cccc|cccc|cc}
\toprule
\multicolumn{1}{l|}{\multirow{2}{*}{Method}} & \multicolumn{4}{c|}{NTU-120 CS}                              & \multicolumn{4}{c|}{NTU-120 CX}                           & \multicolumn{2}{c}{Efficiency} \\ \cmidrule{2-11} 
\multicolumn{1}{c|}{}                        & Joint(\%)         & Bone(\%)          & ~~2s(\%)~~            & ~~4s(\%)~~            & Joint(\%)         & Bone(\%)          & ~~2s(\%)~~            & ~~4s(\%)~~            & FLOPs          & Params        \\ \midrule
STGCN++                 & 92.6          & \textbf{92.6} &  94.0             & 94.3          & \textbf{94.5} & 94.6          & 95.8          & 96.1          & 6.9G           & 1.4M          \\
SkeleT-STGCN++              & \textbf{92.7} & \textbf{92.6}    & \textbf{94.1} & \textbf{94.5} & \textbf{94.5} & \textbf{94.9} & \textbf{95.9} & \textbf{96.3} & \textbf{2.6G}  & \textbf{1.2M} \\ \midrule
CTRGCN                  & \textbf{92.8}   & 92.7           & 94.0         & 94.3             & 94.5          & 94.8          & \textbf{95.9} & \textbf{96.3} & 7.5G           & 1.4M          \\
SkeleT-CTRGCN               & \textbf{92.8} & \textbf{92.9}  & \textbf{94.1}  & \textbf{94.5} & \textbf{94.7} & \textbf{94.9} & \textbf{95.9} & \textbf{96.3} & \textbf{2.5G}  & \textbf{1.1M} \\ \midrule
DGSTGCN                 & 92.6          & \textbf{92.8} & 94.1          & 94.3          & 94.4          & \textbf{95.1} & \textbf{96.0} & 96.1          & 6.3G           & 1.6M          \\ 
SkeleT-DGSTGCN              & \textbf{93.1} & \textbf{92.8} & \textbf{94.3} & \textbf{94.6} & \textbf{94.8} & \textbf{95.1} & \textbf{96.0} & \textbf{96.4} & \textbf{2.4G}  & \textbf{1.3M} \\ \bottomrule
\end{tabular}
}
\label{tab:benchmark-ntu120}
\end{table*}

\begin{table*}
\centering
\vspace{2mm}
\caption{Performance comparison with SOTA multi-modality methods on the NTU-60 and NTU-120 datasets. * indicates using Expressive Keypoints. S, R, and D denote Skeleton, RGB, and Depth.}
\resizebox{0.98\linewidth}{!}{
\begin{tabular}{llcccc}
\toprule
\multirow{2}{*}{Method} & \multirow{2}{*}{Modalities~~~~~~} & \multicolumn{2}{c}{NTU-60} & \multicolumn{2}{c}{NTU-120} \\ \cmidrule{3-6} 
                        &                            & ~~CS(\%)~~       & ~~CV(\%)~~      & ~~CS(\%)~~        & ~~CX(\%)~~       \\ \midrule
STAR-Transformer~\cite{StarTransformer}        & S + R                      & 92.0         & 96.5        & 90.3          & 92.7         \\
VPN++~\cite{vpn++} (w/ 3D Poses)~~~~     & S + R                      & 94.9         & 98.1        & 90.7          & 92.5         \\
HCMFN~\cite{HCMFN}                   & S + R + D~~~~                  & 95.2         & 98.0        & 89.9          & 92.7         \\
MMNet~\cite{MMNet}                   & S + R                      & 96.0         & 98.8        & 92.9          & 94.4         \\
RGBPoseConv3D~\cite{PoseC3D}           & S + R                      &\textbf{97.0} &\textbf{99.6} &\textbf{95.3} &\textbf{96.4}         \\ \midrule
\textbf{Ours: SkeleT}*      & S                          &\textbf{97.0} &\textbf{99.6} & 94.6         &\textbf{96.4}         \\ \bottomrule
\end{tabular}
}
\label{tab:ntu-multi}
\end{table*}

\begin{table}[!]
  \centering
  \begin{minipage}{0.49\linewidth}
  \centering
      \caption{Performance comparison with the SOTA multi-modality methods on PKU-MMD.}
  \resizebox{\linewidth}{!}{
\begin{tabular}{llc}
\toprule
Method & Modalities~ & ~PKU-MMD(\%) \\ \midrule
TSMF~\cite{TSMF}   & S + R     & 95.8      \\
MMNet~\cite{MMNet}  & S + R     & 97.4      \\ \midrule
\textbf{Ours: SkeleT}*~~~ & S         &\textbf{98.4}      \\ \bottomrule
\end{tabular}
    }
    \label{tab:pku-multi}
  \end{minipage}
  \hfill 
  \begin{minipage}{0.47\linewidth}
  \centering
      \caption{Performance comparison with the SOTA multi-modality methods on N-UCLA.}
  \resizebox{\linewidth}{!}{
\begin{tabular}{llc}
\toprule
Method              & Modalities~~~~ & N-UCLA(\%) \\ \midrule
VPN++~\cite{vpn++} & S + R     & 93.5       \\
MMNet~\cite{MMNet}               & S + R     & 93.7       \\ \midrule
\textbf{Ours: SkeleT}*              & S         &\textbf{97.6}       \\ \bottomrule
\end{tabular}
    }
    \label{tab:ucla-multi}
  \end{minipage}
  \vspace{-2mm}
\end{table}
\section{Supplementary experiments}
\subsection{Benchmarking GCN methods on Expressive Keypoints}
\label{supp:benchmarking}
With the fine-grained human body representations provided by Expressive Keypoints, most GCN methods can significantly enhance accuracy by simply adjusting the input keypoints. Our proposed Skeleton Transformation (SkeleT) strategy can be applied to these methods, forming our SkeleT-GCN models, which achieves comparable or even higher accuracy with substantially lower computation cost. 
We conduct a comprehensive benchmark on the NTU-60 and NTU-120 datasets for three representative GCN methods: STGCN++~\cite{PYSKL}, CTRGCN~\cite{CTRGCN}, and DGSTGCN~\cite{DGSTGCN} with Expressive Keypoints as input, as well as their SkeleT-GCN counterparts: SkeleT-STGCN++, SkeleT-CTRGCN, and SkeleT-DGSTGCN. We measure the Top-1 accuracy of joint-stream (Joint), bone-stream (Bone), two-stream fusion (2s)~\cite{2sAGCN}, and four-stream fusion (4s)~\cite{MSAAGCN}. As shown in~Tab.~\ref{tab:benchmark-ntu} and~Tab.~\ref{tab:benchmark-ntu120}, our methods obtain better performance and efficiency than baselines in terms of Top1-accuracy, FLOPs, and number of parameters.

\subsection{Comparison with the state-of-the-art multi-modality methods}
\label{supp:multi-mod-comparison}
Across three benchmarks for skeleton action recognition, including NTU RGB+D \cite{ntu60,ntu120}, PKU-MMD~\cite{pku}, and N-UCLA~\cite{ucla}, our method not only surpasses all skeleton-based methods but also achieves the best performance among all single-modality methods (RGB-based, point cloud-based). To further demonstrate the superiority of strategically employing Expressive Keypoints, we compare our method with previous SOTA multi-modality methods. It can be observed that on the NTU-60 and NTU-120 datasets (Tab.~\ref{tab:ntu-multi}), we achieve comparable performance to the SOTA multi-modality method RGBPoseC3D~\cite{PoseC3D} in three out of four evaluation protocols. On the PKU-MMD dataset (Tab.~\ref{tab:pku-multi}) and the N-UCLA dataset (Tab.~\ref{tab:ucla-multi}), we outperform the SOTA multi-modality method~\cite{MMNet}. 

The experimental results demonstrate that our method, despite being based on a single-modality skeleton input, achieves comparable or even higher performance with a lightweight computational cost than multi-modality methods. This remarkable result primarily stems from introducing fine-grained limb details to the skeleton and employing a SkeleT strategy for effective feature modeling, providing a promising solution for the community.

\begin{figure}[h]
    \centering
    \resizebox{0.99\linewidth}{!}{
        \includegraphics{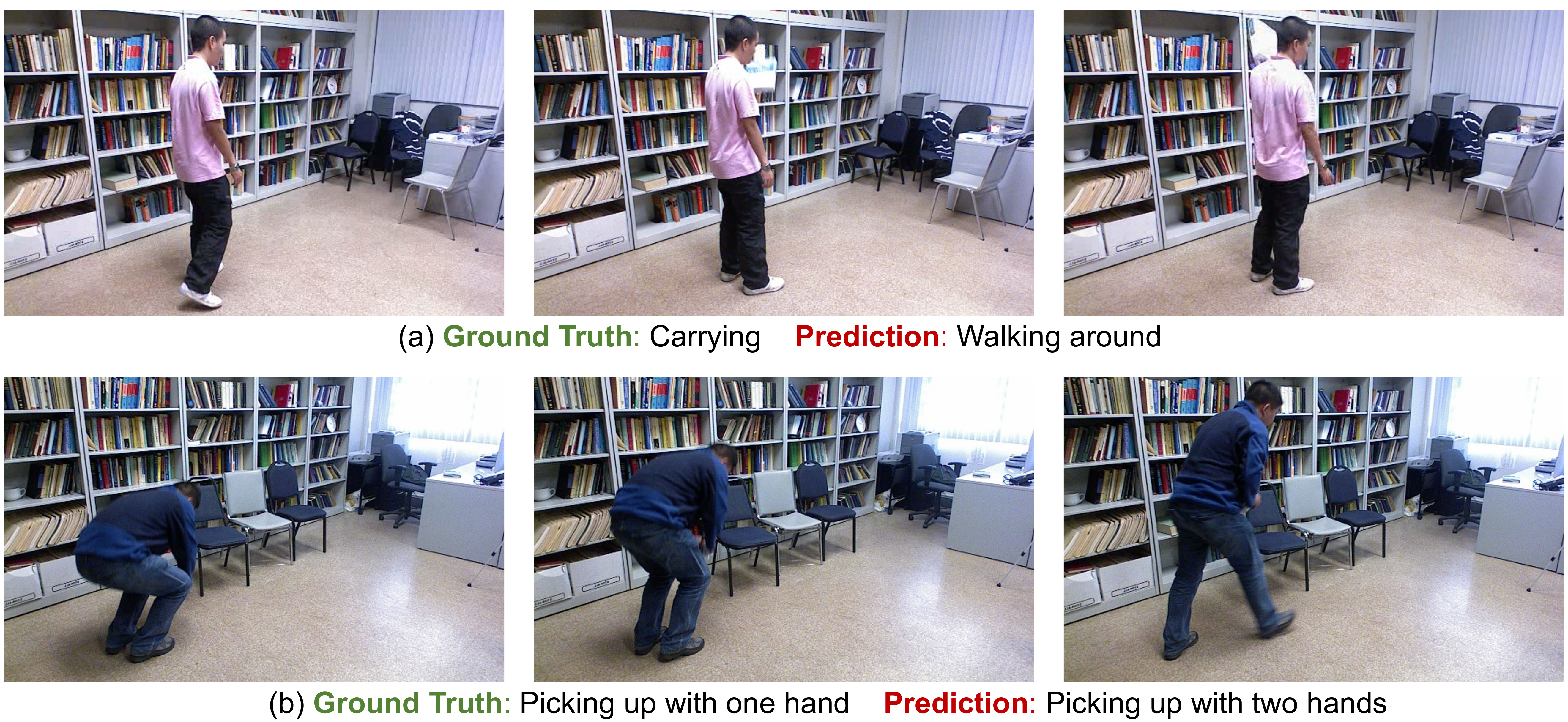}}
    \caption{Failure cases because of lacking depth information on the N-UCLA dataset. \textbf{(a).} The action \textit{carrying} is misclassified as the action \textit{walking around}. \textbf{(b).} The action \textit{picking up with one hand} is misclassified as the action \textit{picking up with two hands}.}
    \label{fig:failure3d}
\end{figure}

\begin{figure}[h]
    \centering
    \resizebox{0.99\linewidth}{!}{
        \includegraphics{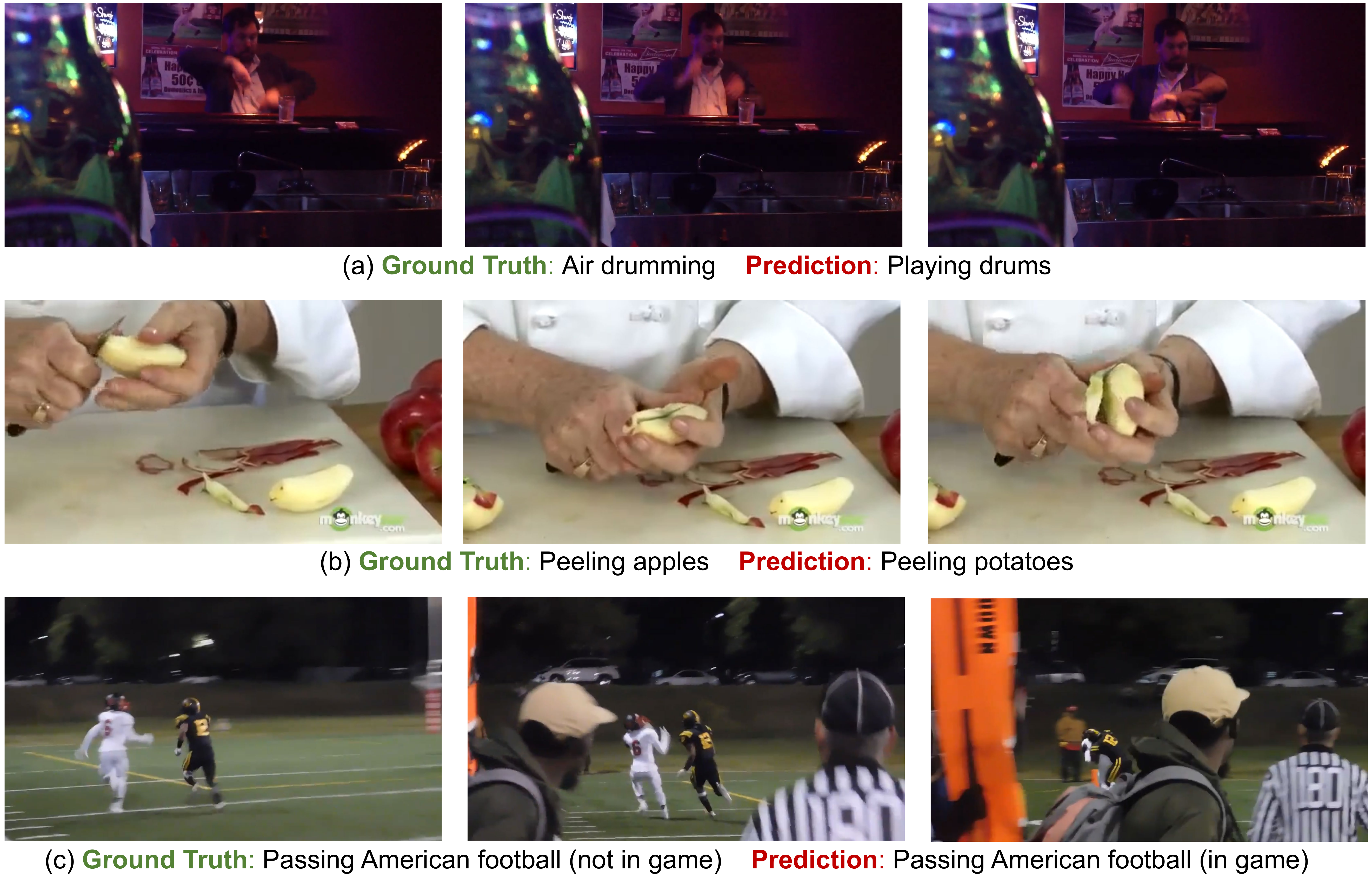}}
    \caption{Failure cases because of lacking object and scene information on the Kinetics-400 dataset. \textbf{(a).} The action \textit{air drumming} is misclassified as the action \textit{playing drums}. \textbf{(b).} The action \textit{peeling apples} is misclassified as the action \textit{peeling potatoes}. \textbf{(c).} The action \textit{passing American football (not in game)} is misclassified as the action \textit{passing American football (in game)}.}
    \label{fig:failureobj}
    \vspace{-5mm}
\end{figure}

\section{Limitations}
\label{supp:Limitations}
\textbf{(i)} Compared to 3D keypoints, our method faces challenges when recognizing actions in occluded scenarios due to the inherent lack of depth information. \textbf{(ii)} Although we extend our method to in-the-wild scenarios using the Instance Pooling module, it still struggles to distinguish certain scene-based actions or human-object interactions due to the lack of capturing of objects and scenes.

\section{Failure cases}
\label{supp:failure-case}

As discussed in Sec.\ref{supp:Limitations}, this section delineates some notable instances where our methodology encounters limitations, leading to classification errors. Specifically, within the N-UCLA dataset, the action labeled as \textit{carrying} is misclassified due to the obstruction of the right hand, which plays a crucial role in the execution of this action, by the body, as depicted in~Fig.~\ref{fig:failure3d}\textcolor{red}{a}. Similarly, Fig.~\ref{fig:failure3d}\textcolor{red}{b} shows \textit{picking up with one hand} is misclassified as \textit{picking up with two hands} because the left hand is completely obscured, making it impossible to distinguish whether the object was picked up with one or both hands. 

Furthermore, on the Kinetics-400 dataset, there are some failure cases shown in~Fig.~\ref{fig:failureobj}\textcolor{red}{a} and~Fig.~\ref{fig:failureobj}\textcolor{red}{b}. The misclassification of those actions are owing to a deficiency in perceiving objects. Moreover, in~Fig.~\ref{fig:failureobj}\textcolor{red}{c}, our method cannot discern the same actions \textit{passing American football} with different context (\textit{in game} vs. \textit{not in game}), stemming from a lack of contextual scene information.

These failure cases reveal that although 2D Expressive Keypoints can significantly enhance recognition performance by providing detailed representations, they struggle in situations involving occlusion due to the absence of depth information, and they cannot effectively distinguish human-object interactions and scene-based actions. These insights point towards promising directions for future enhancements, including the incorporation of depth information and the partial integration of object and scene contextual data.

\section{Social impact}
\label{supp:Impact}
Our research on skeleton-based human action recognition offers significant positive societal impacts, including advancements in healthcare and rehabilitation, elderly care, human-robot interaction, sports analytics, and security and surveillance. However, some potential negative societal impacts may include: (i) the possibility of misuse in surveillance, leading to privacy concerns if individuals are monitored without their consent, and (ii) the risk of biased decision-making if the model is trained on biased data, potentially resulting in unfair treatment of certain groups. However, our model only uses skeletal information, which contains less identifiable appearance information compared to RGB images and videos. This greatly reduces the likelihood of the aforementioned risks.



\newpage

\end{document}